%% file: main.tex
\documentclass[letterpaper]{article}

\PassOptionsToPackage{numbers, compress}{natbib}
\usepackage[main, preprint]{neurips_2026}

\usepackage[utf8]{inputenc} %
\usepackage[T1]{fontenc}    %
\usepackage{hyperref}       %
\usepackage{url}            %
\usepackage{booktabs}       %
\usepackage{amsfonts}       %
\usepackage{nicefrac}       %
\usepackage{microtype}      %
\usepackage{xcolor}         %

\usepackage{enumitem}       %

\usepackage{subcaption} 
\usepackage{graphicx}
\usepackage{makecell}

\definecolor{sns_blue}{HTML}{001C7F}
\definecolor{sns_orange}{HTML}{B1400D}
\definecolor{sns_green}{HTML}{12711C}

\newcommand{\RR}{I\!\!R} %

\usepackage{amsthm, amsmath, amssymb}

\theoremstyle{definition} %
\newtheorem{theorem}{Theorem}

\newtheorem{proposition}[theorem]{Proposition}
\newtheorem{definition}[theorem]{Definition}

\usepackage{algorithm}
\usepackage{algpseudocode}

\usepackage{pifont}
\newcommand{\bestresult}[1]{\textcolor{brown!85!black}{#1}}
\newcommand{\secondresult}[1]{\textcolor{violet!70!black}{#1}}

\newcommand{\sigstar}{\makebox[0.8em][c]{$\star$}}
\newcommand{\nosigstar}{\makebox[0.8em][c]{}}

\usepackage{booktabs}
\usepackage{multirow}
\usepackage[table]{xcolor}
\usepackage{siunitx}
\usepackage{array}
\usepackage{wrapfig}

\newcommand{\deemph}[1]{\textcolor{black!60}{#1}}

\newcommand{\psgnn}{\textsc{PSA-GNN}}
\newcommand{\ipgnn}{IP-GNN}
\newcommand{\poolgnn}{Pool-GNN}

\title{
Hierarchical Graph Representation Learning with Pooling-Induced Substructures

}

\author{%
  Luca Sbicego\thanks{Work conducted while at the University of Oxford.} \\
  EPFL
  \\
  \texttt{luca.sbicego@epfl.ch} \\
   \And
  Xiaowen Dong \\
  University of Oxford \\
  \texttt{xdong@robots.ox.ac.uk} \\
  \And
  Dorina Thanou \\
  EPFL \\
  \texttt{dorina.thanou@epfl.ch} \\
}

\begin{document}

\maketitle

\begin{abstract}
Hierarchical graph-level tasks that require reasoning over substructures that extend beyond local neighborhoods are posing challenges for standard Graph Neural Networks (GNNs). While graph pooling aims to address this by coarsening graphs into hierarchical representations, existing methods often fail to explicitly preserve the substructures they induce, leading to inconsistent empirical performance. Moreover, evaluating such approaches remains challenging due to the lack of benchmarks with explicit hierarchical structure. We propose Pooled-Substructure-Aware GNNs, an approach to graph pooling that explicitly preserves and processes substructures uncovered by structure-based pooling schemes. We show that this design enhances the expressive power of pooling-based GNNs in the sense of the Weisfeiler-Lehman (WL) test. To enable principled evaluation, we introduce two new synthetic datasets with controllable hierarchical structure and propose a metric to characterize the prevalence of hierarchical patterns in real-world graph-level benchmarks. On tasks with relevant hierarchical structure, our approach outperforms matched pooling baselines, while remaining competitive on structurally simpler benchmarks. Code and datasets are publicly available at \url{https://github.com/lsbicego/PSA-GNN/}.

\end{abstract}

\input{sections/introduction}

\input{sections/background}

\input{sections/model_section}

\input{sections/hierarchical}

\input{sections/experiments}

\input{sections/related_works}

\input{sections/conclusion}

\newpage

\begin{ack}

This work received no specific external funding and the authors declare no competing interests.
\end{ack}

\bibliographystyle{unsrtnat}
\bibliography{references}
\newpage

\appendix

\section*{Appendix Overview}
The appendix contains the following sections: 
\begin{itemize}
    \item \textbf{Appendix~\ref{app:proofs}: Proofs for Theoretical Analysis.} We present formal proofs and additional examples for the theoretical analysis presented in Section~\ref{sec:expressivity}.
    \item \textbf{Appendix~\ref{app:brec}: Realized Expressivity on BREC.} We validate our theoretical results from Section~\ref{sec:expressivity} on a benchmark testing model expressivity~\cite{wang2023_brec}.  
    \item \textbf{Appendix~\ref{app:more_synthetic_datasets}: Additional Results on Synthetic Datasets.} We discuss additional results on our synthetic datasets for learned pooling schemes popular in practice and we extend our synthetic datasets with meaningful node features. 
    \item \textbf{Appendix~\ref{app:datasets}: Datasets.} We present details on the datasets used in our experimental evaluation. We first explain the exact generation procedure to generate our synthetic benchmarks. Then, we give a short overview and summary statistics across datasets.
    \item \textbf{Appendix~\ref{app:experiments}: Experimental Details.} We report the necessary details to fully reproduce our experiments. We also briefly discuss the impact of hyperparameters on real-world datasets.
    \item \textbf{Appendix~\ref{app:additional_results}: Results for Additional Subgraph Policies.} We present  and briefly discuss additional results on the synthetic and real-world datasets for three more subgraph policies. These are not included in the main text as these are mainly devised by us, and therefore do not have a strong basis in literature. 
\end{itemize}
\newpage

\input{sections/appendix_proofs}

\input{sections/appendix_BREC}

\input{sections/appendix_learned_pooling}

\input{sections/appendix_experiments}

\end{document}

%% file: sections/introduction.tex
\section{Introduction}

\begin{figure*}[ht]
    \centering
\includegraphics[width=1.0\linewidth,trim=0 15 40 0,clip]{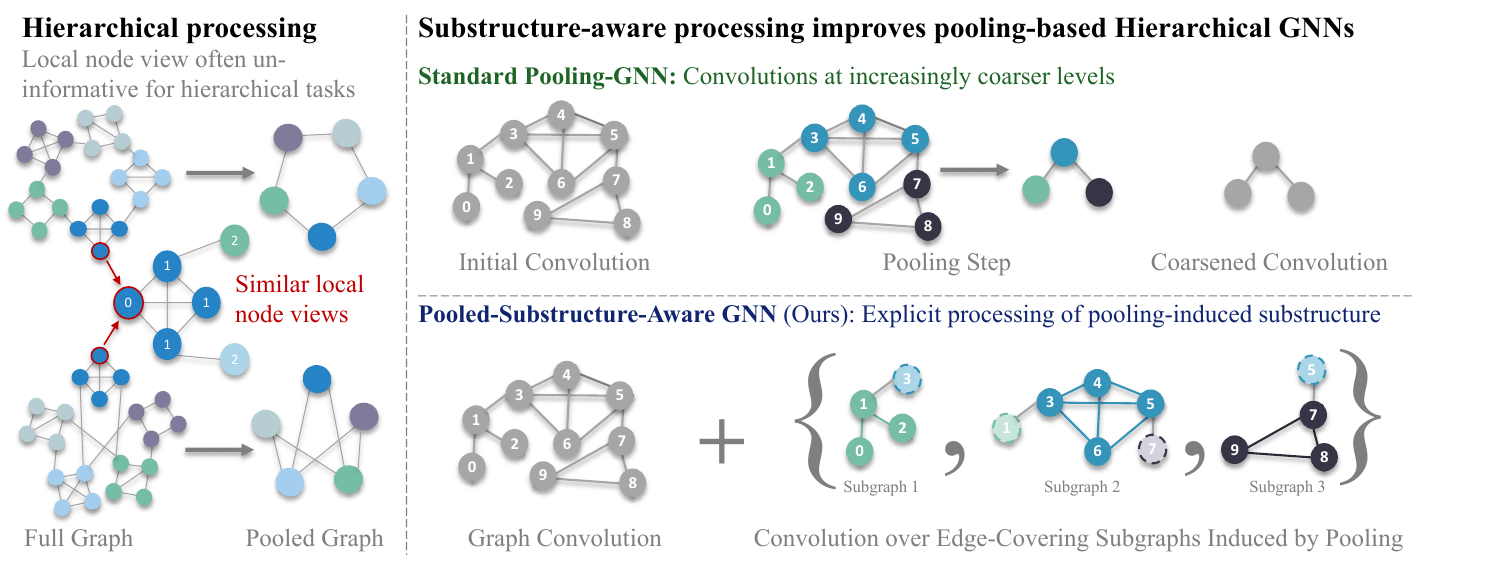}
    \caption{
    \textbf{Motivation and overview.}
    Hierarchical tasks are challenging to solve for node-based MPNNs and GTs. We introduce PSA-GNN, a framework to process meaningful substructures exposed by graph pooling that is more expressive and improves performance on hierarchical graph-level tasks.
    }
    \label{fig:pull_figure}
\vspace{-0.5cm}
\end{figure*}

Graph Neural Networks (GNNs) are widely used to process graph-structured data~\cite{scarselli2008graph, wu2020_comprehensive}, but their prevalent message passing formulation introduces a strong locality bias~\cite{gilmer2017_neural, xu2018_GIN}. Many graph tasks, however, require reasoning over structures extending beyond local neighborhoods, such as communities, or other forms of hierarchical organization~\cite{girvan2002_community_organization, ravasz2003_hierarchical_organization, clauset2008_hierarchical_helps} (Figure~\ref{fig:pull_figure}, left panel).
While Message Passing Neural Networks (MPNNs) and Graph Transformers (GTs) can capture such patterns in principle, they must infer the underlying hierarchical organization implicitly, which can be challenging.
Graph pooling has emerged as a standard mechanism for hierarchical graph representation learning~\cite{liu__2023_pooling_lit_review}, where graph convolutions are interleaved with pooling layers, which coarsen the input graph by grouping nodes into supernodes based on connectivity or learned node features~\cite{ grattarola_understanding_gnn_pooling_2022}.

In principle, this should enable multi-scale reasoning, making graph pooling a natural foundation for hierarchical modeling. However, in practice, pooling often yields limited or inconsistent gains over simpler flat MPNNs~\cite{mesquita_rethinking_2020}. 
Widely used schemes that learn to coarsen a graph based on its node features in an end-to-end fashion have been shown to perform on par with randomized implementations that assign nodes to supernodes by chance \cite{mesquita_rethinking_2020}. Similarly, this type of pooling does not increase the expressivity of the underlying graph convolution in terms of the Weisfeiler-Lehman (WL) test, but at best preserves it \cite{lachi2025_expressive_pooling, bianchi_expressive_2023}. 
In contrast, structure-based pooling schemes that group nodes into hard clusters based on connectivity patterns inaccessible to standard MPNNs have been shown to increase expressivity~\cite{lachi2025_expressive_pooling}. This suggests that graph pooling can expose meaningful higher-order structure beyond simple coarsening of the input graph. Concretely, each supernode can be viewed as exposing a substructure, given by the induced subgraph over its constituent nodes and their internal edges. Yet, existing pooling-based approaches ignore this exposed structure: they collapse each cluster into a single supernode and discard the corresponding subgraph (Figure~\ref{fig:pull_figure}, top right panel).

At the same time, evaluating graph pooling approaches remains a significant challenge. %
Hierarchical GNNs are typically assessed on standard benchmarks, where the extent of hierarchical patterns relevant to the graph-level label is unclear~\cite{wang2024_comprehensive_pooling_benchmark, errica2022_fair_comparison}. Synthetic datasets with controllable hierarchical structure and difficulty are missing, making it difficult to isolate when and why pooling is beneficial.

These observations reveal both an architectural and an empirical gap in hierarchical graph learning: pooling-induced substructures are not explicitly preserved, and existing benchmarks make it difficult to assess when such structure matters. We address these gaps through the following contributions.

\begin{itemize}
    \item \textbf{Novel Framework for Hierarchical Modeling Using Graph Pooling (Section 3.1).}
We introduce Pooled-Substructure-Aware GNN (\psgnn{}; Figure~\ref{fig:pull_figure}, bottom right panel), a principled framework that processes pooling-induced clusters as explicit substructures rather than collapsing them into supernodes, enabling richer hierarchical representations.

\item \textbf{Theoretical Expressivity Analysis (Section 3.2).} We provide a theoretical analysis showing that incorporating pooling-induced substructures increases expressivity in the WL sense compared to standard pooling. 

\item \textbf{Hierarchical Benchmarks (Section 4.1).} We introduce two synthetic datasets, \textsc{Macro-Hierarchies} and \textsc{Granular-Hierarchies}, enabling a controlled evaluation of hierarchical reasoning capabilities. 

\item \textbf{Principled Evaluation.} 
We further present a first characterization of the presence of hierarchical substructures in real-world benchmarks based on Nested Stochastic Block Models (Section 4.2). 
Together with our hierarchical benchmarks, this leads to a clear interpretability of empirical performances: 
on synthetic benchmarks and real-world datasets exhibiting hierarchical structure, our approach improves over existing pooling baselines, while remaining competitive on structurally simpler benchmarks (Section 5).

\end{itemize}

%% file: sections/background.tex
\section{Background}

\textbf{Notation.} Let an undirected, unweighted graph be denoted by $G=(V,E)$, where $V$ and $E$ are the node and edge sets, respectively, and $|V|=N$. We define the adjacency matrix $\mathbf{A}\in\{0,1\}^{N \times N}$ and node features $\mathbf{X}\in\RR^{N \times F}$, where $F$ is the feature dimension. When comparing the expressivity of models, we use the following definition. 

\begin{definition}[Expressivity Comparisons, following \cite{lachi2025_expressive_pooling}]
\label{def:expressiveness_comparison}
Let $\theta, \phi$ be two permutation-invariant functions mapping graphs to vectors. Let $G, H$ be any two graphs. Then:
\begin{itemize}
    \itemsep-0.1em
    \item $\theta$ is \emph{at least as expressive as} $\phi$ if $\phi(G) \neq \phi(H) \implies \theta(G) \neq \theta(H)$.
    \item $\theta$ and $\phi$ are \emph{equally expressive} if $\phi(G) \neq \phi(H) \Longleftrightarrow \theta(G) \neq \theta(H)$.
    \item $\theta$ is \emph{more expressive than} $\phi$ if $\theta$ is at least as expressive as $\phi$ and there exists a pair of graphs $G, H$ such that $\theta(G) \neq \theta(H)$ but $\phi(G) = \phi(H)$.
\end{itemize}
\end{definition}

\textbf{Graph pooling.} A graph pooling operator maps a graph $G$ to a coarsened graph $G'$ with $|V'| < |V|$. We adopt the standard \textsc{Sel-Red-Con} view of pooling \cite{grattarola_understanding_gnn_pooling_2022}. Nodes are first assigned to supernodes via \textsc{Sel}, node features are then aggregated within each supernode by \textsc{Red}, and connectivity between supernodes is determined by \textsc{Con}. Hierarchical GNNs interleave such pooling steps with message passing in order to process a graph at increasingly coarser levels.

\textbf{Graph pooling in theory and practice.} Graph pooling has often delivered inconsistent empirical gains, with hierarchical architectures frequently failing to outperform simpler flat baselines \cite{mesquita_rethinking_2020}. 
Many popular pooling operators are node-based, with assignments or scores learned from the same local node representations produced by the underlying MPNN. 
Recent expressivity results show that such pooling schemes cannot exceed the expressive power of the base encoder when their assignments depend only on these node representations \cite{lachi2025_expressive_pooling, bianchi_expressive_2023}. This may help explain why learned pooling can fail to introduce new structural information. These observations motivate us to investigate pooling from both theoretical and benchmarking perspectives. 

\textbf{Structure-based pooling.} By contrast, recent work has shown that pooling schemes determining cluster assignments based on graph structure rather than learned node features \emph{can} increase expressivity \cite{lachi2025_expressive_pooling}. 
In this setting, pooling-induced clusters can be viewed not only as supernodes for coarsening, but also as structurally meaningful subgraphs of the original graph. This perspective forms the basis of our approach. %

\textbf{Subgraph GNNs.} 
Subgraph GNN methods \cite{bevilacqua2021_ESAN, zhao2021_gnnak, zhang2021_nestedGNN, frasca2022_sun} aim to improve graph representations by processing a graph through a collection of overlapping subgraphs rather than only through the full graph itself. Intuitively, this allows nodes to be represented in multiple local contexts, strengthening expressivity beyond standard message passing \cite{bevilacqua2021_ESAN, zhao2021_gnnak}. This raises the question of how to exploit pooling-induced subgraphs.

%% file: sections/model_section.tex
\section{From Graph Pooling to Substructure-Aware Processing
}
\label{sec:model}

We introduce a principled framework for processing \emph{meaningful pooling-induced substructures}, instantiated by Pooled-Substructure-Aware GNNs (\psgnn). 
The key idea is to treat each pooled cluster as an induced subgraph rather than immediately collapsing it into a supernode. 
This contrasts with standard graph pooling, which reduces each cluster to a supernode and continues computation on the coarsened graph, thereby discarding the internal structure induced by the clusters.

\subsection{
Processing Pooling-Induced Substructures}
\label{sec:model_architecture}

Our perspective naturally connects graph pooling to the Subgraph GNN setting, where models operate not only on the input graph itself, but also on a collection of subgraphs extracted from it \cite{bevilacqua2021_ESAN, zhao2021_gnnak, zhang2021_nestedGNN, frasca2022_sun}. 
In our setting, the subgraph collection is not generated by a separate subgraph selection rule, but arises directly from the substructures exposed by the pooling operator: 

\begin{definition}[Pooling-induced subgraph policies]
\label{def:pooled_subgraphs}
Let a graph pooling scheme partition the nodes of $G=(V,E)$ into disjoint clusters $\mathcal{A}_1,\dots,\mathcal{A}_m$. A subgraph policy $\pi$ induces a set of subgraphs $\mathcal{S}=\{G_i=(V_i,E_i)\}_{i=1}^m$ with  $V_i=\mathcal{A}_i$ and $E_i=\{(u,v)\in E : u,v\in V_i\}$.
Note that the induced subgraphs are vertex-disjoint. We further consider a padded policy $\hat{\pi}$ that augments each subgraph with crossing edges and their incident boundary nodes, yielding an edge-covering set of subgraphs.
\end{definition}

\begin{figure*}[t]
    \centering
    \includegraphics[width=1.\linewidth]{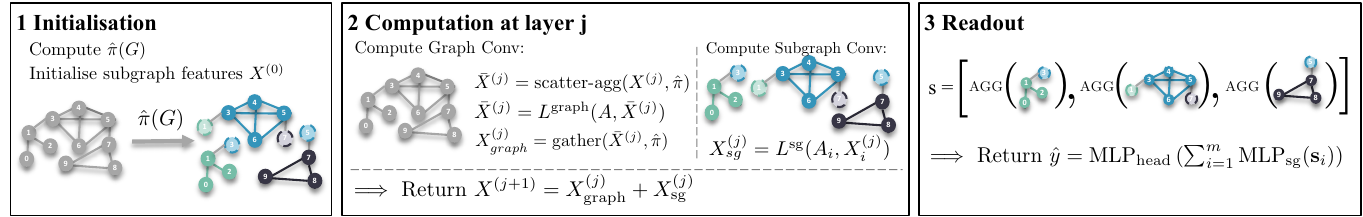}
    \caption{\textbf{Computational flow of PSA-GNN.} 
    We pre-compute the subgraph policy and initialize node features in subgraph space. Each layer then updates these representations by combining message passing on the original graph with message passing inside each induced subgraph. Finally, node representations are aggregated within each subgraph and passed to a DeepSets-style readout.
    }
    \label{fig:psa_gnn}
\vspace{-0.0cm}
\end{figure*}

Building on these pooling-induced subgraph policies, we introduce \psgnn{}, which processes the pooling-induced subgraphs created by a policy $\pi$ using a subgraph-processing layer inspired by DSS-GNN \cite{bevilacqua2021_ESAN}. The layer combines subgraph-level updates with a graph-level context update, as illustrated in Figure~\ref{fig:psa_gnn}.
For each subgraph, a shared encoder $L^{\mathrm{sg}}$ updates node representations using only the internal structure of that subgraph. In parallel, subgraph-node features are scattered back to the original node space, processed on the original graph by $L^{\mathrm{graph}}$, and then mapped back to each subgraph. This graph-level branch allows information to flow across different pooled subgraphs through the original graph adjacency.

Formally, for each subgraph $i \in \{1,\dots,m\}$, layer $L$ updates the node feature matrix as
\begin{equation}
\mathbf{X}'_i
=
L(\mathbf{A}, \mathbf{X}, \pi)_i =
L^{\mathrm{sg}}(\mathbf{A}_i,\mathbf{X}_i)
+
\bigl(L^{\mathrm{graph}}(\mathbf{A},\bar{\mathbf{X}})\bigr)_{\pi_i},
\label{eq:dss_gnn}
\end{equation}

where $\mathbf{A}_i$ and $\mathbf{X}_i$ denote the adjacency and node feature matrices of subgraph $i$, and $\mathbf{A}$ is the adjacency matrix of the original graph. The matrix $\bar{\mathbf{X}}$ is obtained by scatter-aggregating all subgraph-node features back to the original node space. 
For vertex-disjoint subgraphs, this is simply a re-indexing step; for overlapping subgraphs, duplicate occurrences are averaged. 
The operator $(\cdot)_{\pi_i}$ denotes gathering original-node features to the nodes in subgraph $i$.

For graph-level prediction, we first pool the node features of each subgraph into a subgraph representation $\mathbf{s}_i$, and then apply a DeepSets-style readout \cite{zaheer2017_deepsets}:
\begin{equation}
\hat{y}
=
\operatorname{MLP}_{\mathrm{head}}
\left(
\sum_{i=1}^{m}\operatorname{MLP}_{\mathrm{sg}}(\mathbf{s}_i)
\right).
\label{eq:readout}
\end{equation}
Alternatively, the element-wise transformation $\operatorname{MLP}_{\mathrm{sg}}$ can be replaced by a GNN operating over the pooled graph: $\operatorname{GNN}_{\mathrm{sg}}(\mathbf{A}',\mathbf{Z})$, where $\mathbf{Z}_i=\mathbf{s}_i$ and $\mathbf{A}'$ is the adjacency matrix of the coarsened graph.

Unless otherwise specified, \psgnn{} uses a DeepSets-style readout and the padded variant from Definition~\ref{def:pooled_subgraphs}, which augments each subgraph with crossing edges and their associated boundary nodes. 
This introduces controlled overlap between neighboring pooled substructures, allowing local subgraph updates to access boundary context while the graph-level branch couples information through the original graph adjacency.
To distinguish original cluster nodes from added boundary nodes, we append a binary node feature indicating whether a node belongs to the core cluster or was introduced through padding.

\subsection{Expressivity of Explicit Substructure Processing}
\label{sec:expressivity}
In this section, we establish an expressivity hierarchy that characterizes the benefits of explicitly processing pooling-induced substructures. In particular, we investigate how within-subgraph message passing and information exchange across substructures contribute to the expressive power of pooling-based GNNs.

To this end, we consider two comparison architectures. First, we introduce \ipgnn\ (Intra-Pooling-GNN), a simpler variant that uses the base policy $\pi$ to produce disjoint clusters and extends the standard \textsc{Sel-Red-Con} pooling pipeline by performing message passing within each pooling-induced subgraph before the \textsc{Red} and \textsc{Con} steps. Second, we consider a standard \textsc{Sel-Red-Con} \poolgnn\ as a matched pooling baseline.

We first analyze the effect of within-subgraph message passing. Proposition~\ref{prop:intra_vs_pooling} formalizes the basic intuition that immediate coarsening can discard information already present in the substructures induced by pooling. Even a single message-passing step restricted to these subgraphs can therefore strengthen pooling.

\begin{proposition}
\label{prop:intra_vs_pooling}
Let $\mathcal{C}={\mathcal{A}_1,\dots,\mathcal{A}_m}$ be a fixed partition of the node set into disjoint clusters. Then, \ipgnn\ is at least as expressive as \poolgnn\ in the WL sense. Moreover, there exists a partition $\widetilde{\mathcal{C}}$ for which \ipgnn\ is strictly more expressive than \poolgnn.
\end{proposition}

We next show that \psgnn{} subsumes the simpler within-subgraph construction and can go strictly beyond it once pooled subgraphs are allowed to overlap through padding. Intuitively, this overlap enables the model to couple information across neighboring pooled substructures while preserving the underlying cluster structure identified by pooling.

\begin{proposition}
\label{prop:psgnn_hierarchy}
For any subgraph policy $\pi$, \psgnn\ with supergraph readout is at least as expressive as \ipgnn\ under the same policy. Moreover, there exists a policy $\pi$ such that \psgnn\ using the corresponding padded, edge-covering policy $\hat{\pi}$ together with appropriate node markings and any injective readout is strictly more expressive than \ipgnn\ under $\pi$.
\end{proposition}

Propositions~\ref{prop:intra_vs_pooling} and \ref{prop:psgnn_hierarchy} in combination yield the hierarchy between \psgnn, \ipgnn, and \poolgnn. Full formal statements, proofs, and additional examples are deferred to Appendix~\ref{app:proofs}. Finally, we verify this relationship empirically on BREC, a standard expressivity benchmark in Appendix~\ref{app:brec}.

%% file: sections/hierarchical.tex
\section{Characterizing Hierarchical Graph Structure}
\label{sec:hierarchical}

The results of Section~\ref{sec:model} suggest that explicit substructure processing can increase expressivity by leveraging graph pooling more effectively. We expect these gains to translate into improved downstream performance particularly when graph-level tasks depend on hierarchically organized substructures in the input graph rather than local node-level patterns. 
Figure~\ref{fig:pull_figure} (left panel) illustrates this in an example: two graphs may exhibit identical local neighborhoods around selected nodes while differing in their global community organization. In such situations, local node-level views computed by bounded-depth MPNNs may be uninformative. Instead, GTs can enrich node-level representations with positional or structural encodings, but these mechanisms are not designed to explicitly identify task-relevant structures. Thus, while both model families can potentially capture higher-order patterns, they must learn to identify the relevant structures implicitly, whereas substructure-aware models can process them directly.

However, graph pooling methods are typically evaluated on standard graph-level benchmarks, where the relevance of hierarchical structure to the prediction task is often unclear~\cite{mesquita_rethinking_2020, sbicego2025role, wang2024_comprehensive_pooling_benchmark}.
To address this limitation, we adopt two complementary evaluation strategies: controlled synthetic benchmarks with explicit hierarchical signals, and a structural characterization of real-world datasets.

\subsection{Synthetic Hierarchical Benchmarks}
\label{sec:synthetic_datasets}

To study hierarchical graph modeling in a controlled setting, we introduce two synthetic benchmarks in which graph labels explicitly depend  on hierarchical structure. The datasets are generated using SBMs \cite{holland1983_sbm}; full construction details are deferred to Appendix~\ref{app:synthetic_datasets}. By design, the resulting prediction tasks do not impose an expressivity barrier and remain solvable, in principle, by sufficiently deep MPNNs.

\begin{figure}[ht]
  \centering
  \captionsetup[subfigure]{justification=centering,singlelinecheck=false}

  \begin{minipage}[t]{0.48\linewidth}
    \centering
    \textbf{\textsc{Macro-Hierarchies}}\par\smallskip

    \begin{subfigure}[t]{0.48\linewidth}
      \centering
      \includegraphics[width=\linewidth,trim=25 25 25 25,clip]{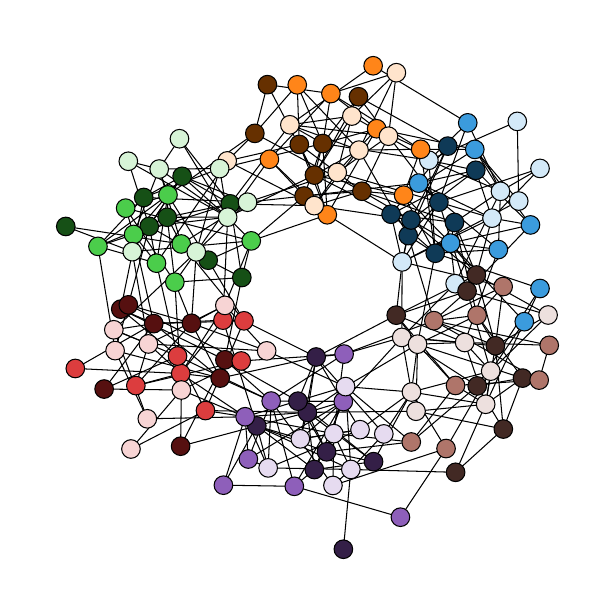}
      \caption{Ring}
      \label{fig:macro-hier-a}
    \end{subfigure}\hfill
    \begin{subfigure}[t]{0.48\linewidth}
      \centering
      \includegraphics[width=\linewidth,trim=25 25 25 25,clip]{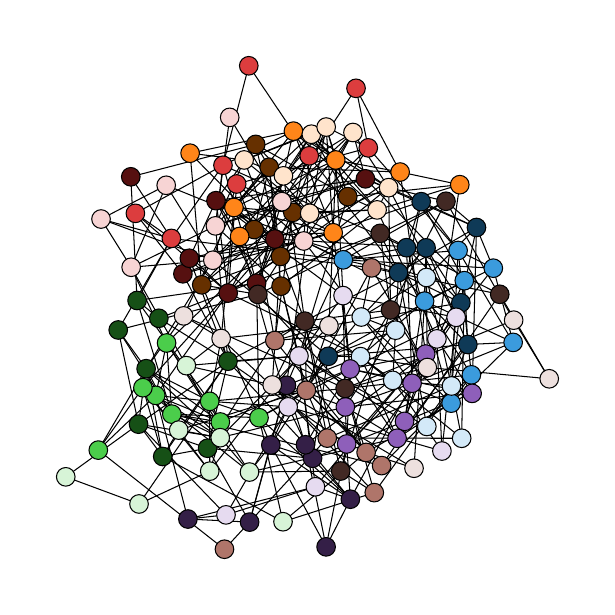}
      \caption{Bipartite}
      \label{fig:macro-hier-b}
    \end{subfigure}
  \end{minipage}
  \hfill
  \begin{minipage}[t]{0.48\linewidth}
    \centering
    \textbf{\textsc{Granular-Hierarchies}}\par\smallskip

    \begin{subfigure}[t]{0.48\linewidth}
      \centering
      \includegraphics[width=\linewidth,trim=25 25 25 25,clip]{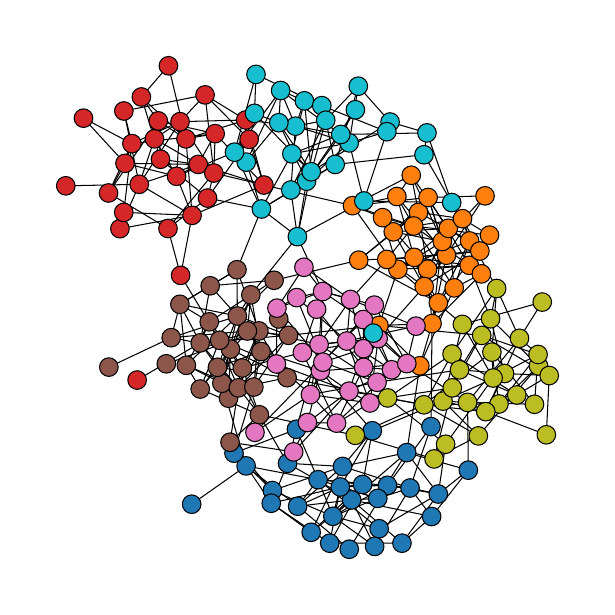}
      \caption{Even Clusters}
      \label{fig:granularity-a}
    \end{subfigure}\hfill
    \begin{subfigure}[t]{0.48\linewidth}
      \centering
      \includegraphics[width=\linewidth,trim=25 25 25 25,clip]{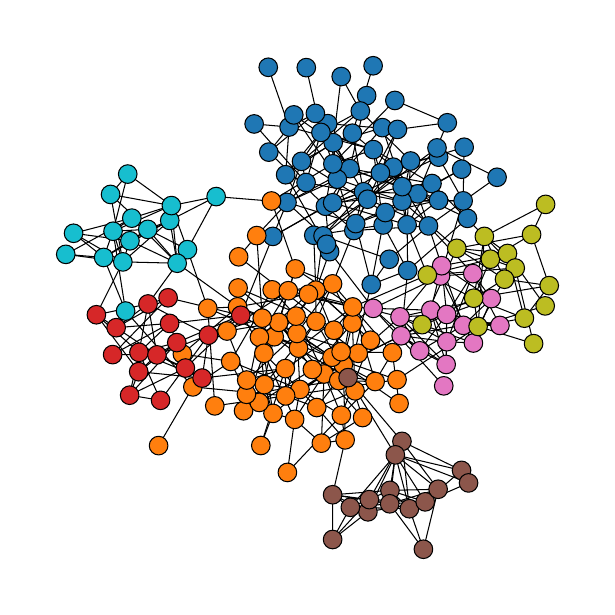}
      \caption{Uneven Clusters}
      \label{fig:granularity-b}
    \end{subfigure}
  \end{minipage}

  \caption{\textbf{Examples from two synthetic datasets.} Left: \textsc{Macro-Hierarchies}, where the class distinction is given by ring versus bipartite higher-level structure. Right: \textsc{Granular-Hierarchies}, where the class distinction is given by even versus uneven cluster sizes.}
  \label{fig:proposed_synthetic_datasets}
\end{figure}

\textsc{Macro-Hierarchies} tests whether a model can reason about higher-level connectivity between clusters, inspired by the example in Figure~\ref{fig:pull_figure}. Each graph is generated from a two-level SBM, with micro-communities grouped into macro-communities. The class signal is encoded at the macro level: macro-communities are arranged either in a ring or in a bipartite pattern, as illustrated in Figure~\ref{fig:proposed_synthetic_datasets}. Distinguishing the two classes therefore requires distinguishing different interaction patterns between larger substructures. We control task difficulty by adding random inter-cluster edges that obscure the underlying ring or bipartite pattern. 

\textsc{Granular-Hierarchies} complements the above dataset and tests whether a model can reason about the size distribution within clusters themselves. Graphs with class $y=0$ consist of equally sized clusters, whereas graphs with class $y=1$ contain a mixture of small and large clusters, as illustrated in Figure~\ref{fig:proposed_synthetic_datasets}. 
The signal therefore lies in the relative granularity of the substructures.
As in \textsc{Macro-Hierarchies},  we can control the difficulty of this task by adding more inter-cluster edges and thus making the substructure signal noisier.

\subsection{Structural Characterization of Real-World Benchmarks}
\label{sec:dataset_analysis}

Complementing our controlled synthetic benchmarks, we characterize the extent of hierarchical structure in real-world graph-level datasets. This provides a structural perspective for interpreting when explicit substructure processing may be beneficial in practice.

For each graph, we fit a degree-corrected nested SBM to its undirected, unweighted topology \cite{holland1983_sbm, karrer2011_degree_sbm, peixoto2014_hierarchicalsbm}, from which we extract two descriptors: the detected \emph{substructure count}, i.e., the number of blocks at the finest inferred level, and the \emph{substructure depth}, i.e., the number of inferred hierarchy levels. Together, these descriptors quantify how much partitionable substructure and hierarchical organization the graph exhibits under the fitted model.

\begin{figure*}[ht]
    \centering
    \includegraphics[width=1.\linewidth]{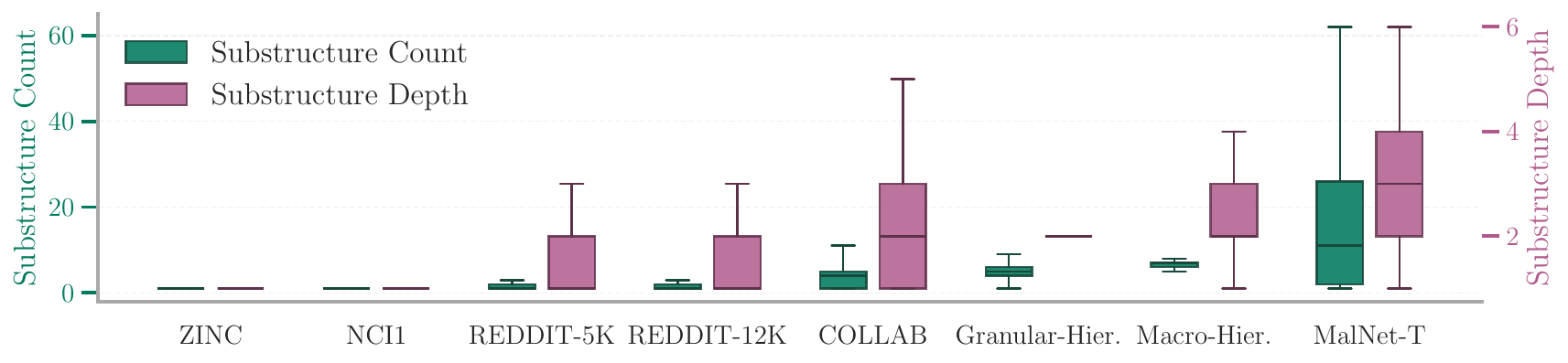}
    \caption{\textbf{Structural characterization of real-world benchmarks using nested SBM fits.} For each graph, we fit a degree-corrected nested SBM and record the number of detected substructures at the finest inferred level (left axis) and the detected substructure depth (right axis). 
    }
    \label{fig:sbm_fits}
\vspace{-0.0cm}
\end{figure*}

Figure~\ref{fig:sbm_fits} shows the distribution of these descriptors across graphs from several widely used graph-level benchmarks. Under this structural probe, ZINC \cite{gomez2018_zinc} and NCI1 \cite{morris2020_tudatasets} appear structurally shallow, as almost all graphs are assigned a single detected substructure at depth~1. In contrast, REDDIT-5K, REDDIT-12K, COLLAB \cite{morris2020_tudatasets}, and MalNet-Tiny \cite{freitas2021largescaledatabasegraphrepresentation} exhibit substantially richer inferred group structure and, for many graphs, multiple hierarchy levels. We use these findings to distinguish structurally flatter datasets from more structure-dominated benchmarks in the experiments below.
To validate the methodology and to provide a point of reference, we also include an instance of our synthetic datasets at a medium difficulty level. As expected, the fitted models recover clear hierarchical structure in these graphs.

%% file: sections/experiments.tex
\vspace{-5mm}
\section{Experiments}
\label{sec:experiments}

We evaluate whether the expressivity gains from explicit substructure processing translate into improved graph-level performance when hierarchical structure is relevant to the prediction task. In all experiments, we compare \psgnn\ to the simple within-subgraph baseline (\ipgnn) as well as the standard pooling baseline (\poolgnn) which immediately coarsens the graph, using matched pooling policies.

\paragraph{Setup.}
Our comparisons are centered on three structure-based pooling schemes: Leiden \cite{traag2019_louvain}, a community-detection method; CurvPool \cite{sanders2023_curvpool}, which clusters nodes using balanced Forman curvature; and CliquePool \cite{luzhnica2019_cliquepool}, which forms clusters from maximal cliques. 
For each policy, we evaluate the three pooling-based architectures for a single pooling step under matched depth and parameter budgets. Throughout all experiments, we use GIN \cite{xu2018_GIN} as the 1-WL base message passing layer. Hyperparameters are chosen on validation data, and the same dataset-specific training protocol is used across models. Full details and implementation choices are provided in Appendix~\ref{app:experiments}. Additional results for exploratory pooling-based subgraph policies are provided in Appendix~\ref{app:additional_results}. 

\subsection{Controlled Hierarchical Benchmarks}
\label{sec:synthetic_results}

\begin{figure}[ht]
  \centering
\begin{minipage}[t]{0.03\linewidth}
  \centering
  \rotatebox{90}{\small\textbf{\quad\quad\quad  \textsc{Macro-Hier.}}}
\end{minipage}\hfill
\begin{minipage}[t]{0.95\linewidth}
\includegraphics[width=1.0\linewidth]{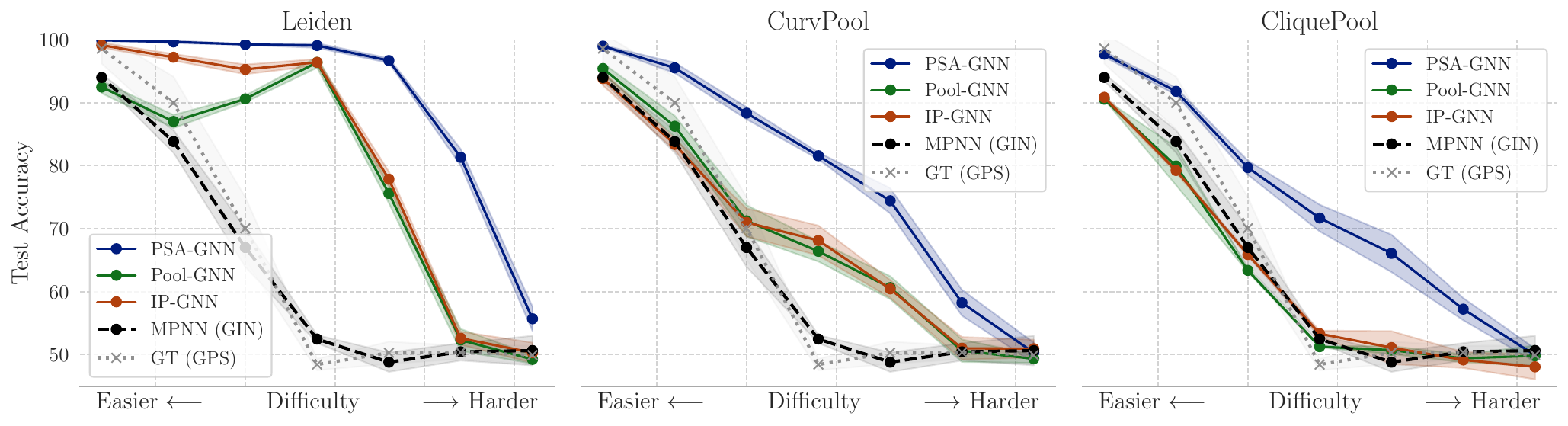}
  \end{minipage}

\begin{minipage}[t]{0.03\linewidth}
  \centering
  \rotatebox{90}{\small\textbf{\quad\quad  \textsc{Granular-Hier.}}}
\end{minipage}\hfill
\begin{minipage}[t]{0.95\linewidth}
\includegraphics[width=1.0\linewidth]{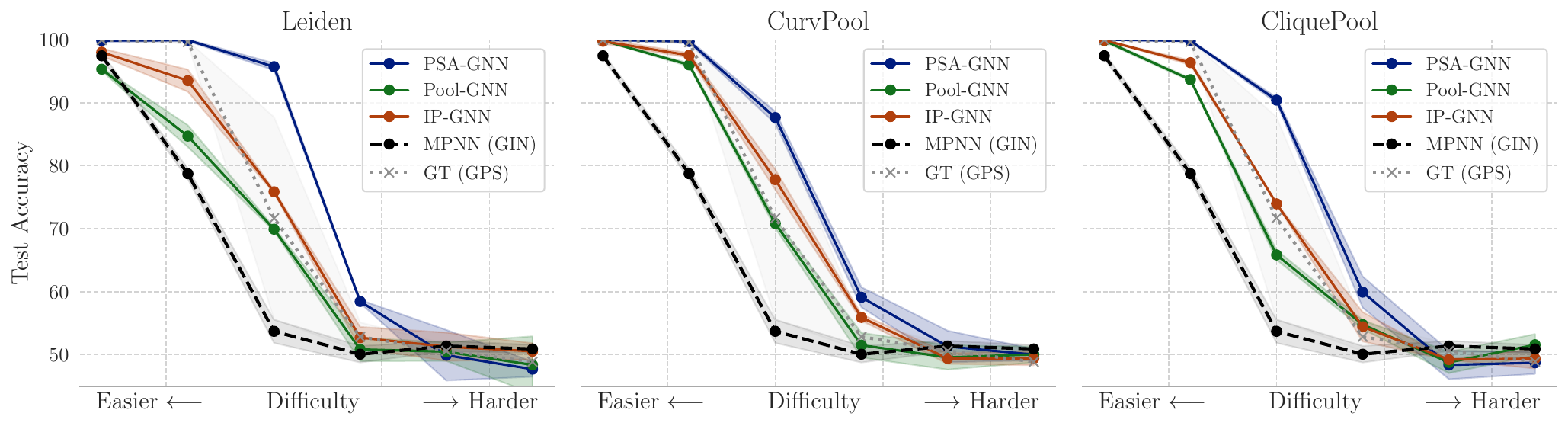}
  \end{minipage}
  
\caption{\textbf{Results on the synthetic hierarchical benchmarks.} We compare \poolgnn, \ipgnn, and \psgnn\ against flat MPNN and GT baselines across representative structure-based schemes. }
  \label{fig:synthetic_results}
\end{figure}

The synthetic benchmarks introduced in Section~\ref{sec:synthetic_datasets} provide a controlled setting for testing our hypothesis, as graph labels  explicitly depend on hierarchical graph structure.
In addition to the matched pooling variants \poolgnn, \ipgnn, and \psgnn, we compare against a flat MPNN based on GIN~\cite{xu2018_GIN} and a GT baseline based on GPS~\cite{rampavsek2022_GPS}. 
Figure~\ref{fig:synthetic_results} reports test accuracy across Leiden, CurvPool, and CliquePool across varying difficulty levels.

Across both synthetic benchmarks, \psgnn\ achieves the best performance across the evaluated difficulty levels and remains substantially more robust as the hierarchical signal becomes harder to recover. The flat MPNN and GPS baselines often degrade more rapidly as difficulty increases, 
supporting the motivation that node-level models can struggle to infer the relevant higher-order organization implicitly. 

The matched pooling variants further show the value of preserving the structures identified by pooling for tasks that depend on higher-order graph organization: \ipgnn\ often improves over immediate coarsening, while \psgnn\ achieves the strongest performance by processing pooling-induced substructures end-to-end. This pattern is consistent with the expressivity hierarchy established in Section~\ref{sec:expressivity} and provides controlled evidence that explicit substructure processing is useful when the task depends on hierarchical graph organization. The magnitude of these gains varies across pooling policies, highlighting that the benefits also depend on how well the resulting substructures align with the task-relevant hierarchy.

We additionally evaluate common learned pooling operators on the synthetic benchmarks in Appendix~\ref{app:hier_gnn_synthetic}. Sparse methods such as SAGPool and ASAPool \cite{lee2019_sagpool, ranjan2020_asap} can perform strongly on \textsc{Macro-Hierarchies}, while dense methods like MinCutPool or DiffPool \cite{bianchi_spectral_2020, ying_hierarchical_2018} do not improve over baselines. These results highlight the diagnostic value of the benchmarks and suggest that different pooling mechanisms capture different hierarchical signals. We further assess the impact of informative node features by repeating the medium-difficulty experiments with contextual SBM features \cite{deshpande2018_contextualSBM} in Appendix~\ref{app:contextual_sbm}. Informative features narrow the performance gap when directly relevant to the task, while \psgnn{} retains its advantage on higher-level structural reasoning.

\subsection{Real-World Benchmarks
}
\label{sec:real_world_results}

Table~\ref{tab:overall_final_march_results} presents results on the real-world benchmarks characterized in Section~\ref{sec:dataset_analysis}. We again focus on the matched comparison between \psgnn, \ipgnn, and \poolgnn\ under the same three structure-based pooling policies. We additionally include popular hierarchical GNN baselines, namely ASAPool, SAGPool, MinCutPool, and DiffPool, as well as a flat MPNN baseline. Where computationally feasible, we also evaluate a Subgraph GNN baseline that applies \psgnn\ to all 2-hop egonets. This provides a reference for subgraph processing with a generic, non-pooling-based subgraph policy, while the matched pooling variants remain the focus of our architectural comparison.

\begin{table}[ht]
\centering
\caption{\textbf{Results on real-world benchmarks.}
Best overall result per dataset is highlighted in
\bestresult{\textbf{brown}}, second best in
\secondresult{\textbf{purple}}. Among \psgnn, \ipgnn, and
\poolgnn, the best method for a given subgraph policy $\pi$ is shown
\textbf{bolded}. A $\star$ indicates that \psgnn{} significantly
outperforms the matched \poolgnn{} under a two-sided Welch test
($p<0.05$, unadjusted).
The \emph{Avg. Rank} reports the average rank of each architecture
level across matched dataset--policy comparisons among \psgnn,
\ipgnn, and \poolgnn. Dense pooling schemes do not scale and
ASAPool cannot use edge features by default.
}
\label{tab:overall_final_march_results}
\setlength{\tabcolsep}{5pt}
\resizebox{\linewidth}{!}{%
\begin{tabular}{@{} l l c c c c c c c @{}}
\toprule
& & & \multicolumn{4}{c}{\textbf{Hierarchical Structure}}
& \multicolumn{2}{c}{\textbf{Flat Structure}} \\
\cmidrule(lr){4-7}\cmidrule(lr){8-9}

\textbf{Model} &
\shortstack[c]{\textbf{Subgraph}\\ \textbf{Policy}} &
\shortstack[c]{\textbf{Avg.}\\ \textbf{Rank} ($\downarrow$)} &
\shortstack[c]{\textbf{COLLAB}\\ \textit{\footnotesize Acc. [\%] ($\uparrow$)}} &
\shortstack[c]{\textbf{REDDIT-5K}\\ \textit{\footnotesize Acc. [\%] ($\uparrow$)}} &
\shortstack[c]{\textbf{REDDIT-12K}\\ \textit{\footnotesize Acc. [\%] ($\uparrow$)}} &
\shortstack[c]{\textbf{MalNet}\\ \textit{\footnotesize Acc. [\%] ($\uparrow$)}} &
\shortstack[c]{\textbf{NCI1}\\ \textit{\footnotesize Acc. [\%] ($\uparrow$)}} &
\shortstack[c]{\textbf{ZINC}\\ \textit{\footnotesize MAE ($\downarrow$)}} \\

\midrule[0.75pt]

\textbf{\psgnn}
& CliquePool
&
& \bestresult{\textbf{78.1 $\pm$ 1.3}}\sigstar
& \secondresult{\textbf{55.0 $\pm$ 0.8}}\nosigstar
& 47.8 $\pm$ 1.4\nosigstar
& 90.6 $\pm$ 2.4\nosigstar
& \secondresult{\textbf{81.5 $\pm$ 2.5}}\nosigstar
& \textbf{0.182 $\pm$ 0.008}\nosigstar \\

& CurvPool
& \textbf{1.36}
& 76.8 $\pm$ 1.5\sigstar
& \textbf{54.9 $\pm$ 1.7}\nosigstar
& \bestresult{\textbf{50.1 $\pm$ 1.7}}\sigstar
& \bestresult{\textbf{92.0 $\pm$ 1.9}}\nosigstar
& 81.2 $\pm$ 1.6\nosigstar
& \textbf{0.177 $\pm$ 0.005}\nosigstar \\

& Leiden
&
& \textbf{76.4 $\pm$ 2.2}\nosigstar
& \bestresult{\textbf{55.3 $\pm$ 2.1}}\sigstar
& \secondresult{\textbf{50.0 $\pm$ 1.3}}\sigstar
& \textbf{91.1 $\pm$ 1.4}\sigstar
& 80.0 $\pm$ 2.5\nosigstar
& 0.213 $\pm$ 0.007\sigstar \\

\midrule[0.75pt]

\textbf{\ipgnn}
& CliquePool
&
& 75.0 $\pm$ 2.3\nosigstar
& 53.3 $\pm$ 3.4\nosigstar
& 45.1 $\pm$ 2.1\nosigstar
& \secondresult{\textbf{91.7 $\pm$ 1.7}}\nosigstar
& 80.6 $\pm$ 2.1\nosigstar
& 0.183 $\pm$ 0.004\nosigstar \\

& CurvPool
& 2.33
& \secondresult{\textbf{77.4 $\pm$ 0.9}}\nosigstar
& 52.6 $\pm$ 3.3\nosigstar
& 42.0 $\pm$ 2.5\nosigstar
& 89.3 $\pm$ 2.0\nosigstar
& 80.3 $\pm$ 1.9\nosigstar
& 0.184 $\pm$ 0.005\nosigstar \\

& Leiden
&
& 73.6 $\pm$ 3.1\nosigstar
& 52.2 $\pm$ 2.6\nosigstar
& 45.6 $\pm$ 0.8\nosigstar
& 90.7 $\pm$ 1.1\nosigstar
& \textbf{80.1 $\pm$ 2.6}\nosigstar
& \textbf{0.186 $\pm$ 0.004}\nosigstar \\

\midrule[0.75pt]

\textbf{\poolgnn}
& CliquePool
&
& 70.4 $\pm$ 4.1\nosigstar
& 52.4 $\pm$ 2.9\nosigstar
& \textbf{48.3 $\pm$ 1.7}\nosigstar
& 90.2 $\pm$ 1.6\nosigstar
& 81.0 $\pm$ 2.4\nosigstar
& \textbf{0.182 $\pm$ 0.004}\nosigstar \\

& CurvPool
& 2.31
& 71.1 $\pm$ 4.5\nosigstar
& 53.6 $\pm$ 2.6\nosigstar
& 47.4 $\pm$ 0.4\nosigstar
& 91.3 $\pm$ 1.1\nosigstar
& \textbf{81.3 $\pm$ 2.0}\nosigstar
& 0.178 $\pm$ 0.004\nosigstar \\

& Leiden
&
& 74.1 $\pm$ 2.3\nosigstar
& 53.1 $\pm$ 1.9\nosigstar
& 44.7 $\pm$ 1.6\nosigstar
& 87.9 $\pm$ 1.5\nosigstar
& 79.9 $\pm$ 2.0\nosigstar
& 0.257 $\pm$ 0.013\nosigstar \\

\midrule[0.75pt]

\deemph{\textbf{Hierarchical GNN}}
& \deemph{ASAPool}
& \deemph{--}
& \deemph{71.6 $\pm$ 5.4\nosigstar}
& \deemph{--}
& \deemph{49.5 $\pm$ 0.2\nosigstar}
& \deemph{88.2 $\pm$ 3.1\nosigstar}
& \deemph{81.2 $\pm$ 2.0\nosigstar}
& \deemph{--} \\

\deemph{}
& \deemph{SAGPool}
& \deemph{--}
& \deemph{70.6 $\pm$ 4.2\nosigstar}
& \deemph{--}
& \deemph{46.6 $\pm$ 1.8\nosigstar}
& \deemph{90.1 $\pm$ 1.1\nosigstar}
& \deemph{80.1 $\pm$ 1.9\nosigstar}
& \deemph{0.416 $\pm$ 0.107\nosigstar} \\

\deemph{}
& \deemph{MinCutPool}
& \deemph{--}
& \deemph{--}
& \deemph{--}
& \deemph{--}
& \deemph{--}
& \deemph{80.2 $\pm$ 2.0\nosigstar}
& \deemph{\secondresult{0.161 $\pm$ 0.003}\nosigstar} \\

\deemph{}
& \deemph{DiffPool}
& \deemph{--}
& \deemph{--}
& \deemph{--}
& \deemph{--}
& \deemph{--}
& \deemph{81.0 $\pm$ 1.7\nosigstar}
& \deemph{\bestresult{0.147 $\pm$ 0.005}\nosigstar} \\

\midrule

\deemph{\textbf{Baselines}}
& \deemph{Egonet}
& \deemph{--}
& \deemph{--}
& \deemph{--}
& \deemph{--}
& \deemph{91.4 $\pm$ 1.4\nosigstar}
& \deemph{\bestresult{81.6 $\pm$ 2.4}\nosigstar}
& \deemph{0.403 $\pm$ 0.060$^{\dagger}$\nosigstar} \\

\deemph{}
& \deemph{MPNN (GIN)}
& \deemph{--}
& \deemph{74.9 $\pm$ 3.7\nosigstar}
& \deemph{53.4 $\pm$ 2.4\nosigstar}
& \deemph{48.3 $\pm$ 2.3\nosigstar}
& \deemph{91.0 $\pm$ 1.6\nosigstar}
& \deemph{80.9 $\pm$ 2.5\nosigstar}
& \deemph{0.172 $\pm$ 0.005\nosigstar} \\

\bottomrule
\end{tabular}%
}

\vspace{2pt}
\begin{minipage}{\linewidth}
\footnotesize
$^{\dagger}$ The Egonet result on ZINC is likely under-converged under the shared training budget.
\end{minipage}
\vspace{-0.5cm}

\end{table}

Across these real-world benchmarks, explicit substructure processing strengthens structure-based pooling. Aggregated over all dataset--policy comparisons, \psgnn\ achieves the best mean rank, ahead of \ipgnn\ and \poolgnn\ (1.36 vs. 2.33 and 2.31 respectively), suggesting that its benefits extend across different pooling policies. These results position \psgnn\ as an effective architecture for real-world hierarchical graph modeling. 

The choice of subgraph policy nevertheless remains task-dependent, as different schemes expose different structural patterns. Since no policy consistently dominates across datasets, domain knowledge or validation performance can guide its selection.

\subsection{Validation of Structural Hierarchy Characterization}

\begin{wrapfigure}[19]{r}{0.35\textwidth}
\vspace{-0.53cm}
    \centering
    \includegraphics[width=0.30\textwidth]{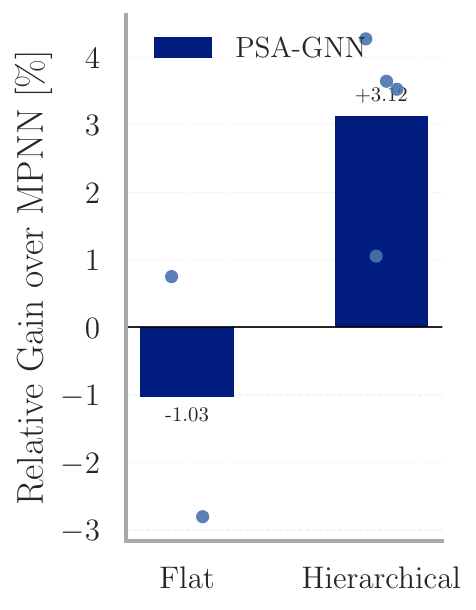}
    \caption{
        \textbf{Relative gain over MPNN.}
        Bars show group means; points show individual datasets.
        Positive values indicate improvement over MPNN.
        }
    \label{fig:regime_gain}
\end{wrapfigure}

Figure~\ref{fig:regime_gain} relates the performance gains of \psgnn\ to the structural characterization introduced in Section~\ref{sec:dataset_analysis}. For each dataset, we report the relative gain of \psgnn\ over the flat MPNN baseline using the best-performing structure-based pooling policy.
In line with our hypothesis, the group-level average gain is positive on benchmarks with richer inferred hierarchical  structure, but slightly negative on structurally flatter datasets. On COLLAB, REDDIT-5K, REDDIT-12K, and MalNet-Tiny, which exhibit richer inferred substructure, \psgnn\ achieves the strongest overall performance, suggesting that explicit substructure processing is particularly beneficial in these settings. On NCI1 and ZINC, \psgnn\ remains competitive, especially within the family of structure-based pooling models, although the overall performance differences are smaller. In these structurally flatter settings, MPNN baselines and pooling schemes with dense soft assignments, such as DiffPool and MinCutPool, can also perform strongly. Overall, these results are consistent with the topology-driven hierarchical structure detected by our SBM probe and support the use of our metric for interpreting when pooling is likely to help.

\begin{wraptable}[12]{l}{0.45\textwidth}
\vspace{-0.0cm}
\centering
\caption{\textbf{Test MAE on GraphBench CO benchmarks.} We compare CO-ER (flat graphs) and CO-RB (with hierarchical structure) on 10K-graph subsets.}

\label{tab:graphbench_er_rb_results}

\setlength{\tabcolsep}{5pt}
\resizebox{\linewidth}{!}{%
\begin{tabular}{@{} l l c c @{}}
\toprule
\textbf{Model} &
\shortstack[c]{\textbf{Subgraph}\\ \textbf{Policy}} &
\shortstack[c]{\textbf{Flat}\\ \textbf{CO-ER}\\ \textit{\footnotesize MAE ($\downarrow$)}} &
\shortstack[c]{\textbf{Hierarchical}\\ \textbf{CO-RB}\\ \textit{\footnotesize MAE ($\downarrow$)}} \\
\midrule[0.75pt]

\psgnn
& CliquePool
& 0.5722 $\pm$ 0.0085
& 1.0108 $\pm$ 0.0145 \\

& CurvPool
& 0.5917 $\pm$ 0.0177
& 0.9345 $\pm$ 0.0649 \\

& Leiden
& 0.6015 $\pm$ 0.0145
& 1.2551 $\pm$ 0.0598 \\

\midrule[0.75pt]

Baseline
& MPNN (GIN)
& 0.5844 $\pm$ 0.0347
& 1.4251 $\pm$ 0.0180 \\

\midrule[0.75pt]

\multicolumn{2}{@{}l}{\textbf{Avg. Gain vs. MPNN}}
& \textbf{-0.70\%}
& \textbf{+25.14\%} \\

\bottomrule
\end{tabular}
}
\end{wraptable}

We further validate our approach on a combinatorial optimization (CO) task of predicting the objective value of a maximum independent set task \cite{stoll2026_graphbenchnextgenerationgraphlearning}. CO-ER \cite{erdos_1959random} consists of flat ER graphs (one substructure across one level), whereas CO-RB \cite{XU_rbgraphs} contains hierarchical graphs according to our structural descriptors, with 22.7 substructures across 4 levels on average.
This contrast clearly illustrates that our structural characterization is indicative of the benefits of explicit substructure processing, with \psgnn{} achieving larger performance gains on datasets exhibiting richer inferred hierarchical structure.

\subsection{Scalability Analysis}
\label{sec:scalability}

Next, we analyze the scalability of \psgnn{} across different subgraph policies. A standard MPNN has time and memory complexity of $\mathcal{O}(N + M)$, where $N$ and $M$ denote the number of nodes and edges, respectively. Excluding subgraph construction, \psgnn\ additionally processes the pooling-induced substructures and thus incurs a policy-dependent overhead factor $c_{\pi}$:

\begin{equation}
    \mathcal{O}\Bigl((1 + c_{\pi})(|V| + |E|)\Bigr), \quad \text{ with } c_{\pi} = \frac{\sum_i (|V_i| + |E_i|)}{N+M}. 
\end{equation}

The value of $c_{\pi}$ depends on the size and overlap of the selected substructures. For example, for a disjoint partition $\pi$ (see Def. \ref{def:pooled_subgraphs}), we have $\sum_i |V_i| = N$ and $\sum_i |E_i| \leq M$, so the subgraph branch adds at most one additional graph-sized computation.

\begin{figure}[ht]
  \centering

  \begin{subfigure}[t]{0.48\linewidth}
    \centering
    \includegraphics[width=1.0\linewidth,trim=0 0 0 0,clip]{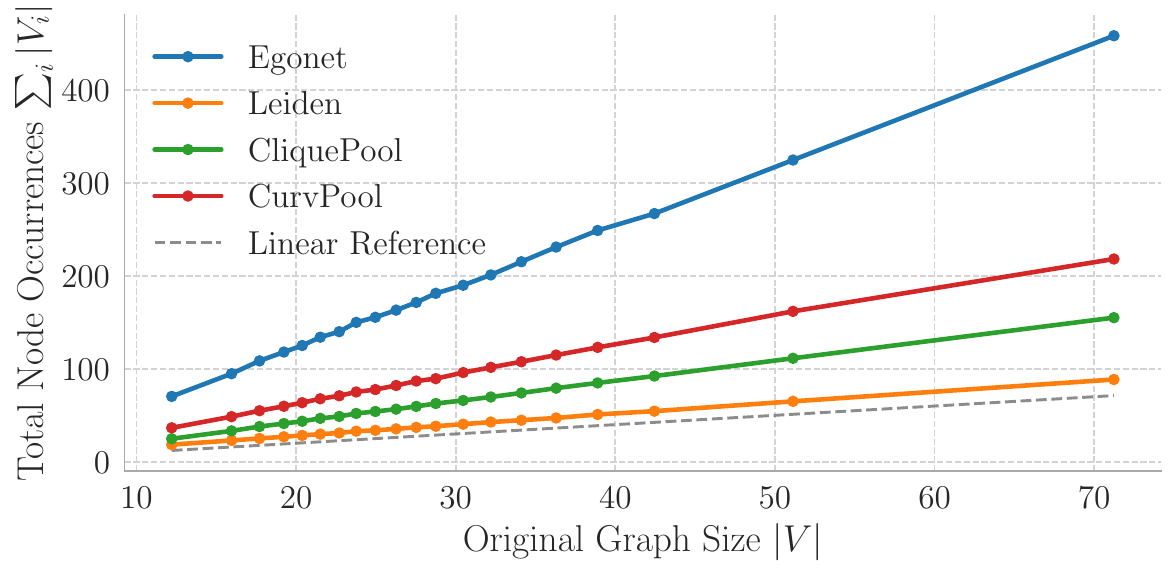}
    \caption{Scalability on \textbf{NCI1}}
    \label{fig:2x2-a}
  \end{subfigure}\hfill
  \begin{subfigure}[t]{0.48\linewidth}
    \centering
    \includegraphics[width=1.0\linewidth,trim=0 0 0 0,clip]{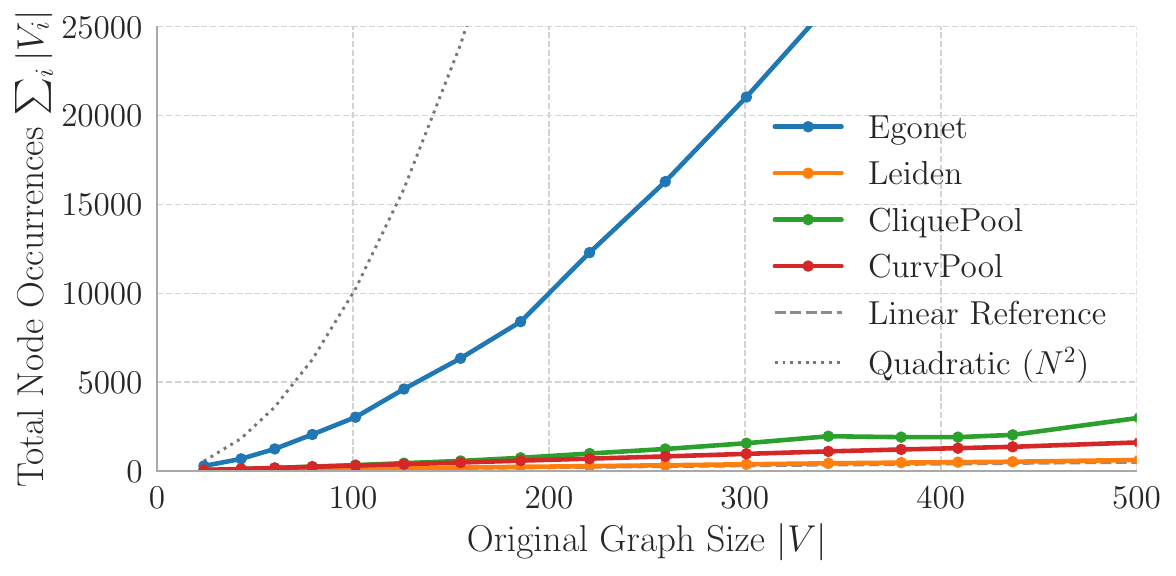}
    \caption{Scalability on \textbf{REDDIT}}
    \label{fig:2x2-b}
  \end{subfigure}\hfill
  \caption{
    \textbf{Scalability of subgraph policies.}
  We compare the total number of subgraph-node occurrences $\sum_i |V_i|$ induced by different policies.
  Observations are grouped into 20 bins and averaged for readability; very large REDDIT graphs are omitted from the plot.
}
  \label{fig:scalability}
\end{figure}

Figure~\ref{fig:scalability} displays the total number of subgraph-node occurrences $\sum_i |V_i|$ induced by different padded policies $\hat{\pi}$ as a function of the original graph size. This quantity drives the factor $c_{\pi}$ and serves as a simple proxy for memory usage during training and inference. 

Pooling-induced policies generate substantially fewer subgraph-node occurrences than generic egonet policies, often by orders of magnitude on larger graphs. 
Processing these substructures therefore remains scalable, making pooled-substructure processing a promising approach for hierarchical modeling on larger graphs. Per-epoch training times are reported in Appendix~\ref{app:experiment_real}.

%% file: sections/related_works.tex
\section{Related Work}

\paragraph{Graph pooling.}
Many widely used pooling architectures~\cite{ liu__2023_pooling_lit_review, grattarola_understanding_gnn_pooling_2022} rely on learned dense assignments or sparse node-selection mechanisms, making learned pooling a central paradigm in hierarchical GNNs.
Motivated by recent results showing that structure-based pooling can increase expressive power~\cite{lachi2025_expressive_pooling}, we revisit simpler pooling policies derived from graph topology.
While some pooling methods use local subgraph operations for scoring or compression~\cite{ranjan2020_asap, xu2026_sshpool}, we propose a different view of structure-based pooling by leveraging the exposed substructures as central objects for hierarchical modeling.

\paragraph{Higher-order Graph Modeling.}
Our work also connects graph pooling to higher-order graph learning approaches that operate on structures beyond individual nodes. In particular, our architecture is inspired by Subgraph GNNs \cite{bevilacqua2021_ESAN, zhao2021_gnnak, zhang2021_nestedGNN, frasca2022_sun}, which improve graph representations by processing collections of subgraphs.
Rather than relying on generic subgraph policies such as egonets or node-deleted subgraphs, we leverage the clusters exposed by graph pooling, directly linking substructure processing to hierarchical graph modeling. This also yields collections of substructures that remain scalable to larger graphs, as is often required in hierarchical modeling. The same distinction applies to graph patching methods such as Graph MLP-Mixer \cite{he2023_mlpmixer}, which tokenize input graphs into node groups. Finally, our work is conceptually related to hypergraph learning \cite{feng2019_hypergraphNN, duta2025_sphinx, kim2024_hypergraph_survey}, where hyperedges can be viewed as meaningful overlapping node groups, providing a useful perspective for future extensions of our framework.

%% file: sections/conclusion.tex
\section{Conclusion}

We introduced a pooled-substructure view of graph pooling by treating induced clusters as valuable substructures for hierarchical tasks. This approach improves hierarchical graph modeling in practice and strengthens pooling from an expressivity perspective. To enable principled evaluation, we proposed two new synthetic datasets with tasks requiring hierarchical reasoning and characterized hierarchical structure in real-world benchmarks.
On tasks with meaningful hierarchical structure, \psgnn\ consistently outperforms immediate coarsening, showing that the substructures exposed by pooling are useful to preserve and process rather than intermediate artifacts to discard. 

At the same time, our experiments clarify the scope of our method. Pooling is beneficial only when graphs exhibit hierarchical structure, and the pooling scheme must expose this structure appropriately, which may vary across datasets. Moreover, our structural characterization should be viewed as a first step toward quantifying hierarchy, as it does not yet assess whether the detected structure is directly predictive of the graph label.

Overall, our results position structure-based pooling combined with explicit substructure processing as a promising direction for hierarchical graph representation learning. Rather than serving solely as a coarsening mechanism, graph pooling can be viewed as a way to expose relevant substructures for computation. This opens a broader design space for hierarchical GNNs and reveals useful connections to other higher-order graph learning frameworks, which can inform future hierarchical modeling architectures.

%% file: sections/appendix_proofs.tex
\section{Proofs for Theoretical Analysis}
\label{app:proofs}

\paragraph{Assumptions.}
Throughout this appendix, we assume that all message passing layers are permutation-equivariant and at most 1-WL expressive, as is standard for MPNNs such as GIN \cite{xu2018_GIN}. We further assume that the within-subgraph message passing layer used by \ipgnn\ can realize the identity map, for instance via residual connections. Finally, whenever pooling reduces node representations to supernode representations, we assume that the reduction operator $RED$ is injective on multisets of node embeddings, so that it does not destroy distinctions already present in the pooled clusters \cite{lachi2025_expressive_pooling, bianchi_expressive_2023}. For \psgnn, we also assume that the final subgraph-level or supergraph-level readout is chosen injectively whenever needed. All expressivity comparisons are understood in the sense of Definition~\ref{def:expressiveness_comparison}.

\subsection{Proof of Proposition~\ref{prop:intra_vs_pooling}}
\label{app:proof_prop_intra}

We show first that \ipgnn\ is at least as expressive as \poolgnn, and then give a graph pair for which it is strictly more expressive.

\paragraph{At least as expressive.}
Let $M_{pool}$ be any \poolgnn\ under a fixed hard partition $\mathcal{C}=\{A_1,\dots,A_m\}$. Consider the corresponding \ipgnn\ under the same partition. By assumption, the additional within-subgraph message passing layer can realize the identity map. Hence we may choose its parameters so that it leaves all node representations unchanged before the subsequent $RED$-$CON$ pooling step. With this choice, \ipgnn\ reproduces exactly the same computation as $M_{pool}$ on every input graph. Therefore, for every \poolgnn\ there exists an \ipgnn\ with identical outputs, and \ipgnn\ is at least as expressive as \poolgnn.

\paragraph{Strictly more expressive.}
It remains to show that there exists a partition $\widetilde{\mathcal{C}}$ and a pair of graphs that can be distinguished by \ipgnn\ but not by \poolgnn. Consider two graphs $G$ and $H$ whose node sets are partitioned into two clusters $A$ and $B$ of equal size, with $G[A]\cong P_4$, $G[B]\cong P_4$, and $H[A]\cong K_{1,3}$, $H[B]\cong K_{1,3}$. Add inter-cluster edges so that both full graphs are regular and so that the coarsened graph obtained after pooling is identical in both cases.

A model in \poolgnn\ immediately reduces each cluster to a supernode. Since the pooled graph is the same for $G$ and $H$, and since the pre-pooling message passing stage cannot distinguish the regular unlabeled input graphs, \poolgnn\ cannot separate the pair.

By contrast, \ipgnn\ performs an additional message passing step inside each induced subgraph before reduction. A 1-WL-bounded MPNN can distinguish $P_4$ from $K_{1,3}$ via their different internal degree structure, and the injective reduction preserves this distinction at the supernode level. Hence \ipgnn\ distinguishes the pair, establishing strict separation. \qed

\begin{figure}[htb]
  \centering
\begin{minipage}[t]{0.08\linewidth}
  \centering
  \rotatebox{90}{\small\textbf{Shrikhande Graph}}
\end{minipage}\hfill
\begin{minipage}[t]{0.90\linewidth}
  \begin{subfigure}[t]{0.23\linewidth}
    \centering
    \includegraphics[width=\linewidth,trim=5 5 5 5,clip]{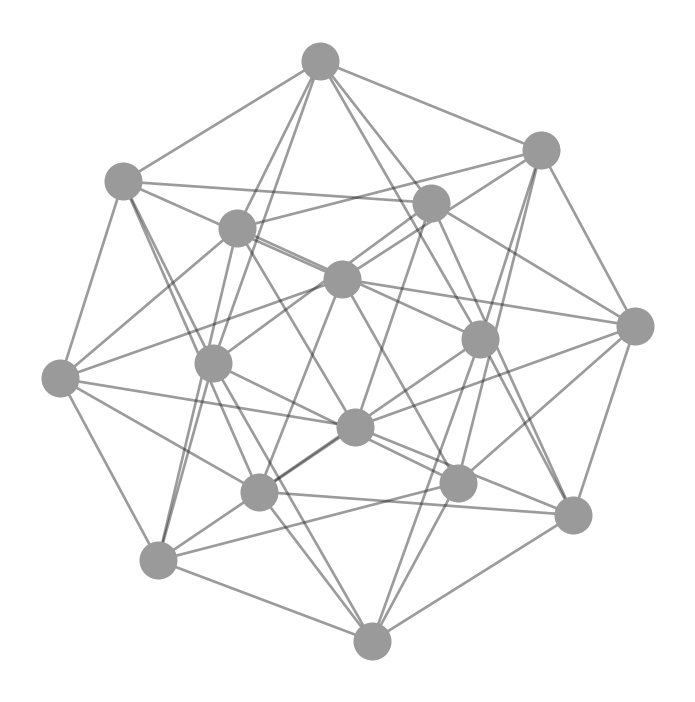}
    \caption{Original graph}
  \end{subfigure}\hfill
  \begin{subfigure}[t]{0.23\linewidth}
    \centering
    \includegraphics[width=\linewidth,trim=5 5 5 5,clip]{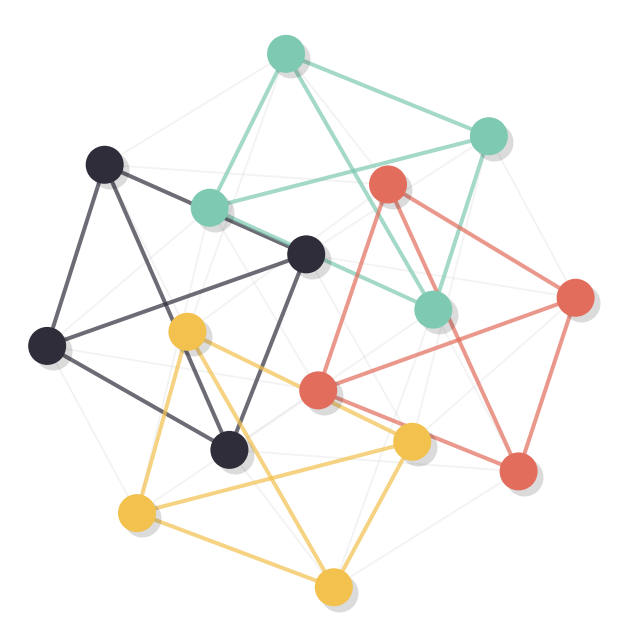}
    \caption{Induced subgraphs}
  \end{subfigure}\hfill
  \begin{subfigure}[t]{0.23\linewidth}
    \centering
    \includegraphics[width=1.1\linewidth,trim=0 0 0 0,clip]{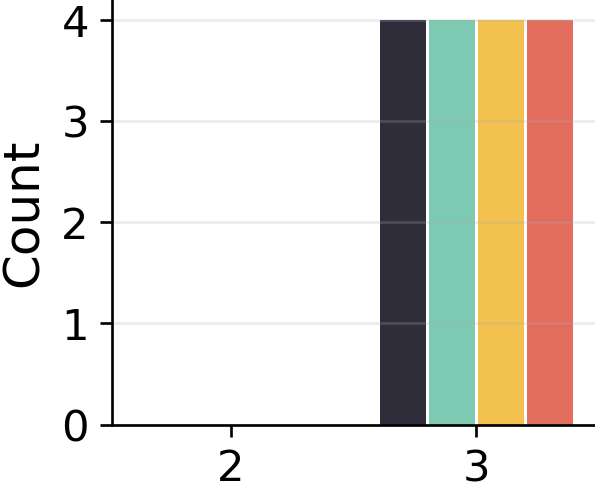}
    \caption{Internal degrees}
  \end{subfigure}\hfill
  \begin{subfigure}[t]{0.23\linewidth}
    \centering
    \includegraphics[width=0.9\linewidth,trim=0 0 0 0,clip]{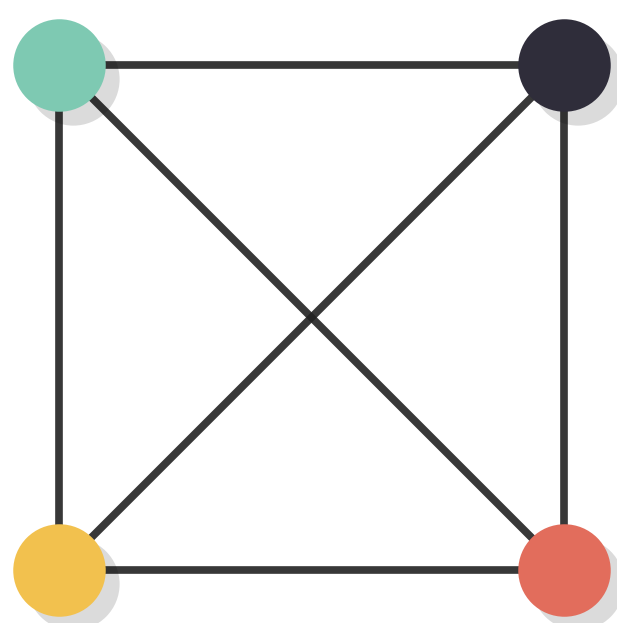}
    \caption{Pooled graph}
  \end{subfigure}
\end{minipage}

\vspace{0.4em}

\begin{minipage}[t]{0.08\linewidth}
  \centering
  \rotatebox{90}{\small\quad\quad\textbf{Rook's Graph}}
\end{minipage}\hfill
\begin{minipage}[t]{0.90\linewidth}
  \begin{subfigure}[t]{0.23\linewidth}
    \centering
    \includegraphics[width=\linewidth,trim=5 5 5 5,clip]{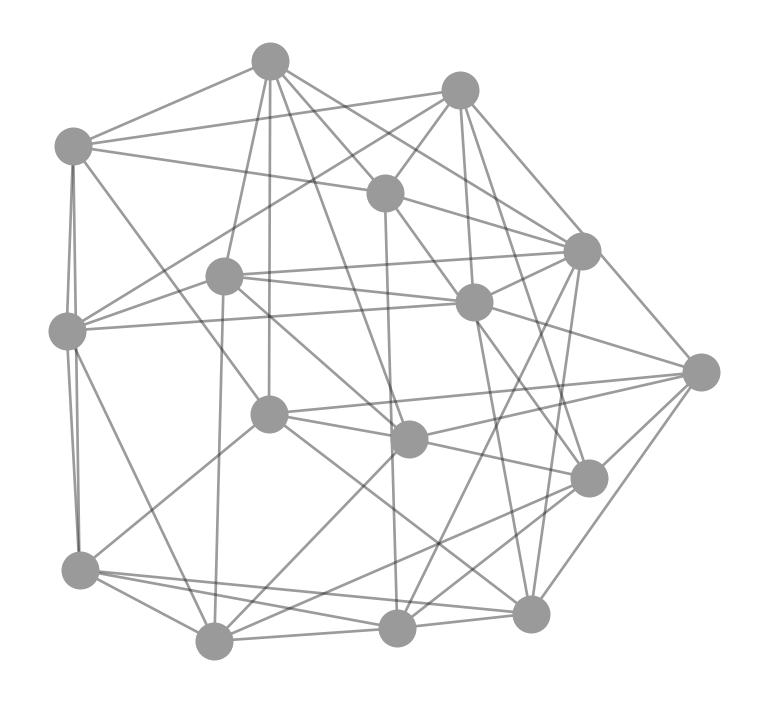}
  \end{subfigure}\hfill
  \begin{subfigure}[t]{0.23\linewidth}
    \centering
    \includegraphics[width=\linewidth,trim=5 5 5 5,clip]{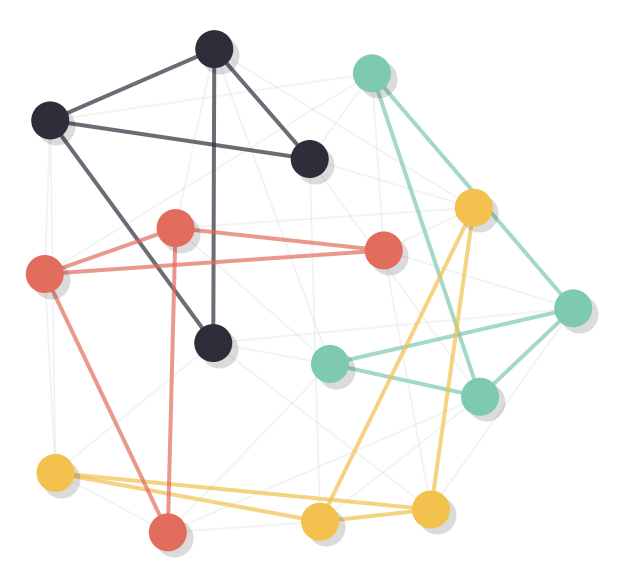}
  \end{subfigure}\hfill
  \begin{subfigure}[t]{0.23\linewidth}
    \centering
    \includegraphics[width=1.1\linewidth,trim=0 0 0 0,clip]{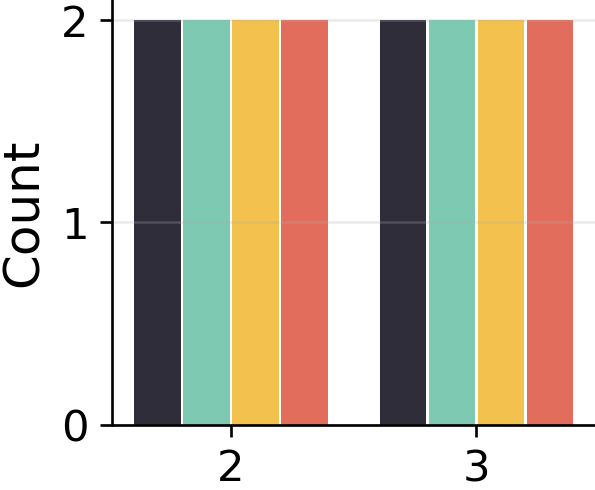}
  \end{subfigure}\hfill
  \begin{subfigure}[t]{0.23\linewidth}
    \centering
    \includegraphics[width=0.9\linewidth,trim=0 0 0 0,clip]{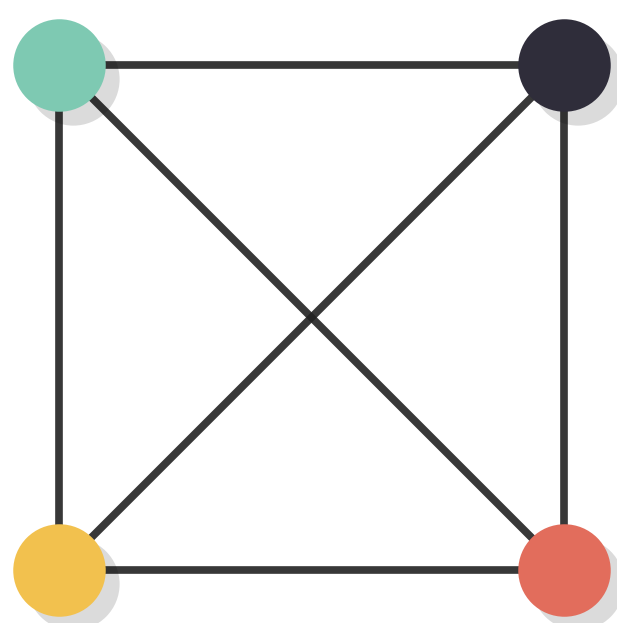}
  \end{subfigure}
\end{minipage}

\caption{Additional example for Proposition~3. The Shrikhande graph and the $4\times4$ Rook graph are not distinguished by immediate coarsening under this Leiden partition, but their pooling-induced subgraphs exhibit different internal degree signatures that can be exploited by \ipgnn\ or \psgnn.}
\label{fig:srg_leiden_appendix}
\end{figure}

\paragraph{Additional example.}
The above argument is not limited to hand-crafted partitions. In practice, structure-based pooling policies can induce subgraphs whose internal structure is informative even when the final coarsened graph is not. Figure~\ref{fig:srg_leiden_appendix} gives an additional example based on the Shrikhande graph and the $4\times 4$ Rook's graph, a classical pair of non-isomorphic strongly regular graphs with parameters $\mathrm{SRG}(16,6,2,2)$ \cite{arvind2020_weisfeiler_examples}. Under an empirical Leiden-based partition, the induced subgraphs expose different internal degree signatures. These are accessible to \ipgnn\ and \psgnn, but are discarded by immediate coarsening.

\subsection{Proof of Proposition~\ref{prop:psgnn_hierarchy}}
\label{app:proof_prop_pooled}

We prove the two claims in turn. First, \psgnn\ with supergraph readout subsumes \ipgnn\ under any fixed policy $\pi$. Second, there exists a policy $\pi$ such that the padded version of \psgnn\ is strictly more expressive than \ipgnn\ under the base policy.

We equip \psgnn{} with a supergraph readout only to match the post-pooling interface of \ipgnn{} in the expressivity argument. In practice, we use a DeepSets-style readout by default, since our hypothesis is that the relevant hierarchical signal is already captured by the processed pooling-induced substructures and their coupling through the original graph branch. 

\paragraph{(1) At least as expressive as \ipgnn.}
Fix any subgraph policy $\pi$. Recall that one \psgnn\ layer updates subgraph-node features as
\begin{equation*}
X_i' \;=\; L^{\mathrm{sg}}(A_i,X_i) \;+\; \bigl(L^{\mathrm{graph}}(A,\bar X)\bigr)_{\pi_i},
\end{equation*}
where $L^{\mathrm{sg}}$ acts inside each selected subgraph and $L^{\mathrm{graph}}$ acts on the original graph after scatter-aggregating subgraph-node features back to node space.

Consider an \ipgnn\ with a single pooling step under the same policy $\pi$. Its computation can be decomposed into three stages:
(i) message passing on the original graph before pooling,
(ii) message passing inside the induced subgraphs before reduction,
and (iii) message passing on the pooled supergraph after $RED$-$CON$.

A \psgnn\ with supergraph readout can emulate these stages sequentially:
\begin{itemize}
    \item Pre-pooling global message passing is recovered by setting $L^{\mathrm{sg}}=0$ and choosing $L^{\mathrm{graph}}$ to match the corresponding original-graph layers.
    \item Within-subgraph message passing is recovered by setting $L^{\mathrm{graph}}=0$ and choosing $L^{\mathrm{sg}}$ to match the intra-subgraph layers.
    \item Since both models use the same policy $\pi$ and the same $RED$-$CON$ mechanism, they induce the same pooled graph. The post-pooling computation of \ipgnn\ can therefore be reproduced by replacing the DeepSets readout of \psgnn\ with a supergraph readout of the same depth.
\end{itemize}
Thus every computation realizable by \ipgnn\ under $\pi$ can also be realized by \psgnn\ under $\pi$. Hence \psgnn\ is always at least as expressive as \ipgnn. 

\begin{figure*}[ht]
    \centering
\includegraphics[width=1.0\linewidth,trim=0 5 10 0,clip]{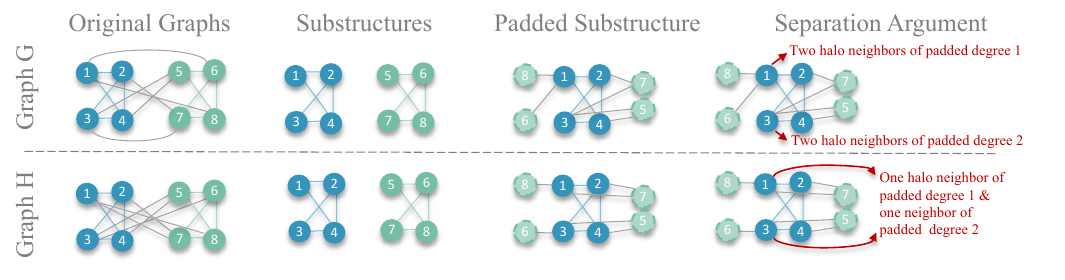}
    \caption{
    Strict-separation example for padded pooled-substructure processing.
    The two original graphs $G$ and $H$ are both $4$-regular and use the same hard partition into two diamond subgraphs.
    Under the unpadded policy $\pi$, both graphs yield the same subgraphs and coarsened graph with six crossing edges, so \ipgnn{} cannot distinguish them.
    Under the padded policy $\hat{\pi}$, however, the $A$-subgraphs expose different crossing-edge incidence patterns.
    In $G$, the core nodes $1$ and $3$ see different halo-neighbor types and split after two $1$-WL refinement steps; while they remain symmetric in $H$.
    }
    \label{fig:padded_proof}
\end{figure*}

\paragraph{(2) Strict improvement under padded policies.}
Having shown that \psgnn{} subsumes \ipgnn{} under $\pi$, we now give an example of a pair of graphs that can be distinguished by \psgnn{} but not \ipgnn{} under a fixed pooling policy. Consider the two 4-regular graphs $G$ and $H$ in Figure~\ref{fig:padded_proof}. Both graphs have vertex set $V=\{1,\ldots,8\}$ and the same hard partition $A=\{1,2,3,4\}, B=\{5,6,7,8\}$, where the induced subgraphs on $A$ and $B$ are identical diamonds $D=K_4-e$. $G$ and $H$ are indistinguishable by 1-WL from constant node features but differ in their crossing edges as shown.

\ipgnn{} cannot distinguish this pair. Under the policy $\pi$, \ipgnn{} sees the same multiset of induced subgraphs, namely $\{D,D\}$, in both graphs. Moreover, the coarsened graph is the same in both cases: two supernodes $A$ and $B$ connected by six crossing edges. Therefore \ipgnn{} cannot distinguish $G$ and $H$ under this policy.

\psgnn{} distinguishes this pair after two computational layers using $\hat{\pi}$ with core/halo node markings. We focus on the padded $A$-subgraph. In both graphs, the core is the same diamond on nodes $\{1,2,3,4\}$, while the halo nodes are $\{5,6,7,8\}$. After one $1$-WL refinement step on the padded subgraph, the halo nodes split into two types: nodes $5$ and $7$ have two core neighbors, while nodes $6$ and $8$ have one core neighbor. In $G$, the two low-internal-degree core nodes receive different halo-neighborhood multisets $1:\{6,8\}$ and $3:\{5,7\}$. Thus node $1$ sees two halo nodes of padded degree $1$, while node $3$ sees two halo nodes of padded degree $2$. In $H$, however, $1:\{8,7\}$ and $3:\{6,5\}$, so nodes $1$ and $3$ each see one halo node of padded degree $1$ and one halo node of padded degree $2$. Consequently, after a second $1$-WL refinement step, nodes $1$ and $3$ receive different colors in the padded $A$-subgraph of $G$, but remain indistinguishable in the padded $A$-subgraph of $H$.

A two-layer subgraph encoder $L^{\mathrm{sg}}$ with a 1-WL-expressive base layer can realize these computations. Together with an injective subgraph readout and an injective permutation-invariant graph-level readout, \psgnn{} distinguishes the padded substructure collections of $G$ and $H$. Hence padded pooled-substructure processing can be strictly more expressive than unpadded intra-pooling processing. \qed

%% file: sections/appendix_BREC.tex
\section{Realized Expressivity on BREC}
\label{app:brec}

\begin{figure*}[ht]
    \centering
    \includegraphics[width=1.\linewidth]{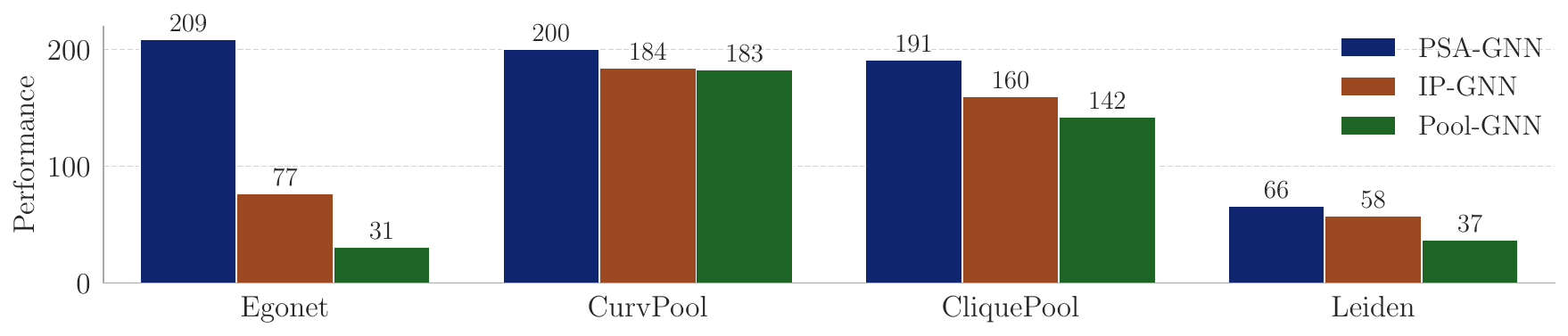}
    \caption{\textbf{Number of BREC pairs distinguished by each architecture-policy combination, out of 400 total.} Following the official BREC protocol, we report the best score achieved over three seeds. }
    \label{fig:brec_results}
    \vspace{-0.3cm}
\end{figure*}

In this section we verify the expressivity hierarchy from Section~\ref{sec:expressivity} on a standard realized expressivity benchmark \cite{wang2023_brec}. We present results for Leiden, CliquePool, and CurvPool as well as the Egonet baseline in Figure~\ref{fig:brec_results}. Clearly, the expected relationship appears in practice: \ipgnn\ improves over immediate coarsening, while \psgnn\ performs best overall. Experimental details are deferred to Appendix~\ref{app:experiments}.

%% file: sections/appendix_learned_pooling.tex
\section{Further Results on Synthetic Datasets}
\label{app:more_synthetic_datasets}

\subsection{Hierarchical GNNs on Hierarchy Benchmarks}
\label{app:hier_gnn_synthetic}
We further evaluated hierarchical GNNs based on popular pooling schemes that learn cluster assignments based on node features. We distinguish between \emph{dense} and \emph{sparse} methods. Dense methods, such as DiffPool and MinCutPool \cite{ying_hierarchical_2018, bianchi_spectral_2020}, typically learn a soft assignment matrix $\mathbf{M} \in \RR^{|V| \times |V'|}$ that maps all input nodes to supernodes. 
\emph{Sparse} pooling methods typically learn to assign a score to each node and retain only the top-$k$ nodes, dropping non-selected nodes. 

We include DiffPool and MinCut as representative dense pooling schemes \cite{ying_hierarchical_2018,bianchi_spectral_2020} as well as TopKPool, SAGPool and ASAPool as sparse variants \cite{gao2019_topk_graph_unets, knyazev2019_top_k_2_understanding, lee2019_sagpool, ranjan2020_asap}. As an additional case study, we include randomized variants of ASAPool and SAGPool \cite{mesquita_rethinking_2020}. 

\begin{figure}[ht]
  \centering
\begin{minipage}[t]{0.03\linewidth}
  \centering
  \rotatebox{90}{\small\textbf{\quad\quad  \textsc{Macro-Hier.}}}
\end{minipage}\hfill
\begin{minipage}[t]{0.95\linewidth}
\includegraphics[width=1.0\linewidth]{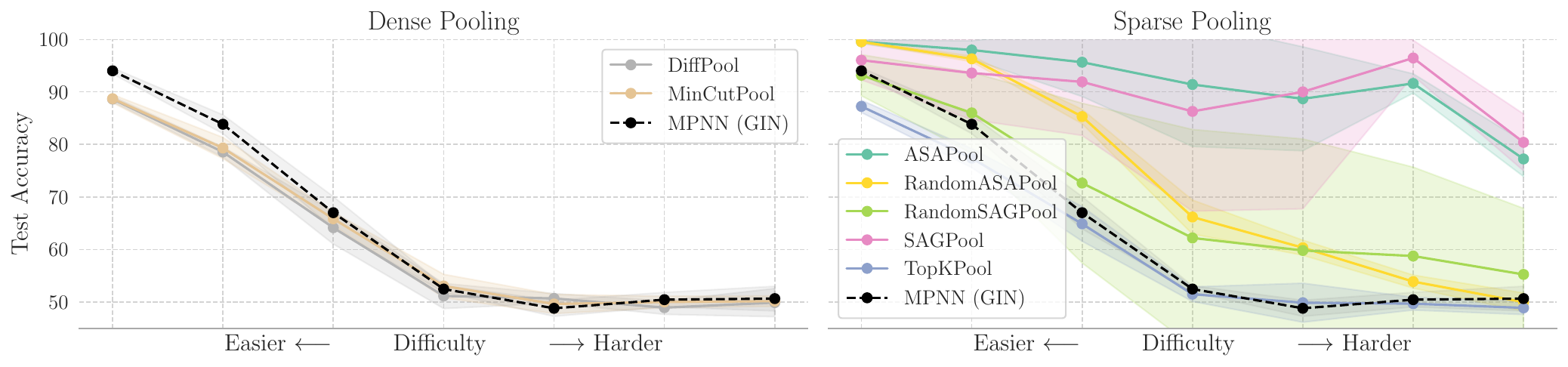}
  \end{minipage}

\begin{minipage}[t]{0.03\linewidth}
  \centering
  \rotatebox{90}{\small\textbf{\quad\quad  \textsc{Granular-Hier.}}}
\end{minipage}\hfill
\begin{minipage}[t]{0.95\linewidth}
\includegraphics[width=1.0\linewidth]{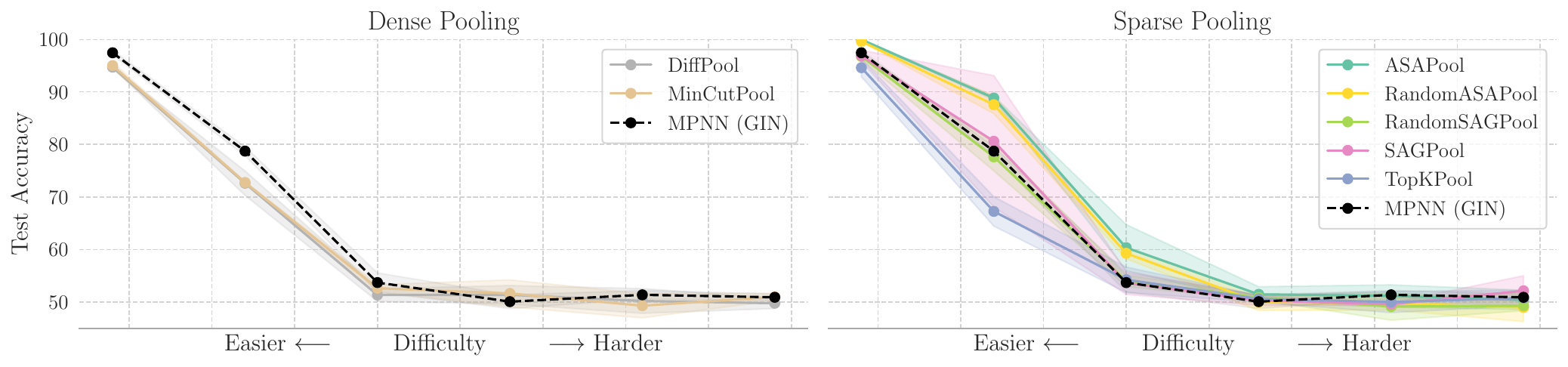}
  \end{minipage}
  
\caption{\textbf{Results on the synthetic hierarchical benchmarks for learned pooling operators.} We compare a range of popular learned pooling operators as well as their randomized implementations.}
  \label{fig:pooling_synthetic_results}
\end{figure}

On \textsc{Macro-Hierarchies}, all dense pooling schemes fail to improve over the MPNN baseline, likely because the pooling step has little informative node-feature signal to exploit. 
By contrast, ASAPool and SAGPool, two sparse pooling baselines, achieve best-in-class performance, especially in the harder regimes, albeit at the cost of a higher variance across seeds. Both also clearly outperform their randomized implementations, indicating that they \emph{learn} meaningful hierarchical structure. Furthermore, TopKPool, which implements a similar but simpler coarsening logic, performs substantially worse, suggesting that success in this setting depends on more than sparsity alone. These findings motivate further research on the promise and failure modes of pooling schemes and validate our synthetic approach as a valuable tool to that end.

On \textsc{Granular-Hierarchies}, all learned pooling-based GNNs fail to beat the MPNN baseline. Similarly to \textsc{Macro-Hierarchies}, dense pooling operators might not learn meaningful cluster assignments from graph topology alone. Sparse schemes on the other hand drop nodes, which might not be a suitable inductive bias as solving the task requires recognizing and estimating cluster sizes. 

\subsection{Contextual Node Features on Hierarchy Benchmarks}
\label{app:contextual_sbm}

The original experiments on our synthetic datasets were conducted with constant node features as these benchmarks are designed to isolate hierarchical
structural reasoning. To test whether this design influences our empirical results, we
repeated the medium-difficulty experiments with $F=16$
node features across varying levels of informativeness. 

Following the contextual SBM
framework~\citep{deshpande2018_contextualSBM} and generalizing the
feature construction in Eq.~9 of \citet{shi2024_homophily_modulation} from binary
labels to multiple communities, we generate
\begin{equation}
\mathbf{x}_i
=
\sqrt{\frac{\mu}{N}}\,\mathbf{u}_{c_i}
+
\boldsymbol{\xi}_i,
\end{equation}

where $c_i$ is the lowest-level community of node $i$, $u_c \sim \mathcal{N}(0, I_F/F)$ are random community prototypes, and $\xi_i \sim \mathcal{N}(0, I_F/F)$.
The prototypes are resampled independently for every graph, so
they provide noisy community information without directly
encoding the graph-level label. Larger $\mu$ corresponds to
more informative features.

\begin{table*}[t]
\centering
\caption{\textbf{Impact of contextual node features on synthetic
hierarchy benchmarks.}
Test accuracy (mean $\pm$ standard deviation)
on the medium-difficulty \textsc{Granular-Hierarchies} and
\textsc{Macro-Hierarchies} datasets with $F=16$ contextual node features.
The gap denotes the accuracy difference in percentage points
between \psgnn{} and the GIN baseline.}
\label{tab:contextual_sbm_results}

\setlength{\tabcolsep}{7pt}
\resizebox{\textwidth}{!}{%
\begin{tabular}{@{} l c c c c c c @{}}
\toprule

& \multicolumn{3}{c}{\textbf{Granular-Hierarchies}}
& \multicolumn{3}{c}{\textbf{Macro-Hierarchies}} \\

\cmidrule(lr){2-4}\cmidrule(lr){5-7}

\textbf{Feature Level}
& \textbf{\psgnn{} (Leiden)}
& \textbf{MPNN (GIN)}
& \textbf{Gap}
& \textbf{\psgnn{} (Leiden)}
& \textbf{MPNN (GIN)}
& \textbf{Gap} \\

\midrule[0.75pt]

Featureless
& 95.70 $\pm$ 0.56
& 53.73 $\pm$ 1.84
& +41.97
& 99.06 $\pm$ 0.36
& 52.50 $\pm$ 0.64
& +46.56 \\

$\mu = 1$
& 95.10 $\pm$ 0.36
& 49.63 $\pm$ 2.15
& +45.47
& 99.37 $\pm$ 0.31
& 49.47 $\pm$ 1.00
& +49.90 \\

$\mu = 5$
& 95.33 $\pm$ 0.47
& 57.07 $\pm$ 2.74
& +38.26
& 99.23 $\pm$ 0.25
& 51.40 $\pm$ 2.49
& +47.83 \\

$\mu = 10$
& 95.97 $\pm$ 0.47
& 66.20 $\pm$ 1.32
& +29.77
& 99.53 $\pm$ 0.06
& 49.60 $\pm$ 2.55
& +49.93 \\

$\mu = 50$
& 97.40 $\pm$ 0.90
& 92.67 $\pm$ 0.60
& +4.73
& 99.40 $\pm$ 0.26
& 50.30 $\pm$ 2.95
& +49.10 \\

\bottomrule
\end{tabular}%
}
\end{table*}

On \textsc{Granular-Hierarchies}, increasingly informative features help GIN identify the lowest-level communities and estimate their sizes, narrowing the performance gap. On \textsc{Macro-Hierarchies}, however, information about these communities alone is insufficient to recover their higher-level ring or bipartite organization, and GIN remains near chance. These results suggest that informative node features can reduce the need for structural reasoning when closely aligned with the prediction target as expected, but do not eliminate \psgnn{}'s advantage when the task requires reasoning about relationships between communities.

%% file: sections/appendix_experiments.tex
\section{Dataset Details}
\label{app:datasets}

Here, we provide additional details on all datasets used in our experiments.

\subsection{Synthetic Datasets}
\label{app:synthetic_datasets}

This appendix section provides details of the exact construction procedure of graphs in both \textsc{Macro-Hierarchies} and \textsc{Granular-Hierarchies}. We generate 10,000 graphs for both benchmarks for various difficulty levels. All nodes are endowed with constant features, unless otherwise specified.

\paragraph{\textsc{Macro-Hierarchies}.}

This dataset is generated by a two-level stochastic block model with $M$ macro-communities, each consisting of $K$ micro-communities of size $S$. The class signal is encoded in the connectivity pattern between macro-communities: at the macro level, they are arranged either as a ring or as a bipartite graph, as illustrated in Figure~\ref{fig:proposed_synthetic_datasets}.

Concretely, we sample $M \in \{6,7,8\}$ and $K \in \{3,4,5\}$, and then choose $S \in \{6,8,10,12\}$ such that $MKS \approx 200$. Nodes within the same micro-community are connected with probability $p_{\text{micro}} = 0.35$, while nodes from different micro-communities within the same macro-community are connected with probability $p_{\text{macro}} = 0.12$. Connections between nodes in different macro-communities are then sampled according to the class pattern (ring or bipartite). If two macro-communities are adjacent in the underlying pattern, edges between them are generated with probability $p_{\text{signal}} = p_{\text{base}} + \delta$, where $p_{\text{base}} = 0.01$ is a small background-noise term and $\delta \in (0.002, 0.014)$ controls the strength of the class signal. For pairs of macro-communities that are not connected in the underlying pattern, edges are sampled with probability $p_{\text{outside}}$. This quantity is chosen so that the expected node degree is matched across labels, preventing trivial separation by simple graph statistics. Varying $\delta$ therefore controls the difficulty of the task.

\paragraph{\textsc{Granular-Hierarchies}.}

In our second dataset, the class signal is encoded in the relative sizes of the clusters. Graphs with $y=0$ consist of equally sized clusters, whereas graphs with $y=1$ contain clusters of unequal size, as illustrated in Figure~\ref{fig:proposed_synthetic_datasets}.

As before, we sample the number of clusters as $M \in \{6,7,8\}$ and fix the total number of nodes to $N = 200$. We further set a target expected degree of $d_{\text{target}} = 5$. For class $y=1$, two clusters are chosen to be four times larger than the remaining ones, whereas for class $y=0$, all clusters have equal size. Edges between nodes in different clusters are sampled with probability $p_{\text{global}} \in (0.002, 0.022)$, which acts as background noise and therefore controls the difficulty of the task. Within-cluster edges are then sampled with probability $p_{\text{within}}$, chosen such that the expected node degree remains close to $d_{\text{target}}$. This prevents trivial separation based on the total number of edges.

\subsection{Dataset Statistics}
\label{app:statistics_datasets}

Table \ref{tab:dataset_summary} presents summary statistics of the main datasets used in our experiments. 

\begin{table*}[ht]
\centering
\caption{\textbf{Summary statistics of the benchmark datasets.} Dataset-level statistics are reported as mean $\pm$ standard deviation over graphs where applicable.}
\label{tab:dataset_summary}
\setlength{\tabcolsep}{5pt}
\resizebox{\textwidth}{!}{%
\begin{tabular}{@{}
l
l
S[table-format=5.0]
S[table-format=2.0]
S[table-format=2.0]
S[table-format=1.0]
c
c
c
@{}}
\toprule
\textbf{Dataset} & \textbf{Task} &
\shortstack[c]{\textbf{\#}\\\textbf{Graphs}} &
\shortstack[c]{\textbf{\# Labels}} &
\shortstack[c]{\textbf{Node feat.}\\\textbf{dim}} &
\shortstack[c]{\textbf{Edge feat.}\\\textbf{dim}} &
\shortstack[c]{\textbf{Avg. \# Nodes}} &
\shortstack[c]{\textbf{Avg. \# Edges}} &
\shortstack[c]{\textbf{Avg.}\\\textbf{Degree}} \\
\midrule[0.75pt]

\multicolumn{9}{@{}l}{\textit{Hierarchical Structure}} \\
COLLAB & Classification & 5000  & 3  & 0  & 0 &
$\num{74} \pm \num{62}$ &
$\num{2457} \pm \num{6439}$ &
$\num{37.4} \pm \num{44.0}$ \\

REDDIT-5K & Classification & 5000 & 5 & 0 & 0 &
$\num{509} \pm \num{453}$ &
$\num{595} \pm \num{567}$ &
$\num{2.3} \pm \num{0.2}$ \\\\

REDDIT-12K & Classification & 11929 & 11 & 0 & 0 &
$\num{391} \pm \num{429}$ &
$\num{457} \pm \num{518}$ &
$\num{2.3} \pm \num{0.2}$ \\

MalNet & Classification & 5000 & 5 & 0 & 0 &
$\num{1522} \pm \num{1650}$ &
$\num{2859} \pm \num{3205}$ &
$\num{3.0} \pm \num{1.2}$ \\

CO-RB & Regression & 10000 & 1 & 0 & 0 &
$\num{241} \pm \num{28}$ &
$\num{4497} \pm \num{1504}$ &
$\num{36.8} \pm \num{9.9}$ \\

\midrule

\multicolumn{9}{@{}l}{\textit{Flat Structure}} \\
NCI1 & Classification & 4110 & 2 & 37 & 0 &
$\num{30} \pm \num{14}$ &
$\num{32} \pm \num{15}$ &
$\num{2.2} \pm \num{0.1}$ \\

ZINC & Regression & 12000 & 1 & 1 & 1 &
$\num{23} \pm \num{4}$ &
$\num{25} \pm \num{5}$ &
$\num{2.1} \pm \num{0.1}$ \\

CO-ER & Regression & 10000 & 1 & 0 & 0 &
$\num{250} \pm \num{29}$ &
$\num{4721} \pm \num{1090}$ &
$\num{37.3} \pm \num{4.4}$ \\

\midrule
\multicolumn{9}{@{}l}{\textit{Synthetic, structure-controlled datasets}} \\
Macro-Hierarchies & Classification & 10000 & 2 & 0 & 0 &
$\num{237} \pm \num{46}$ &
$\num{942} \pm \num{322}$ &
$\num{7.8} \pm \num{1.2}$ \\

Granular-Hierarchies & Classification & 10000 & 2 & 0 & 0 &
$\num{200} \pm \num{0}$ &
$\num{600} \pm \num{23}$ &
$\num{6.0} \pm \num{0.2}$ \\
\bottomrule
\end{tabular}%
}
\end{table*}

Our synthetic datasets are introduced in Section \ref{app:synthetic_datasets}. We briefly describe the real-world benchmarks in the domains of malware analysis, social interaction, scientific collaboration, and molecular property prediction. \textbf{MalNet-Tiny} contains function call graphs of malicious software and requires predicting the malware family, providing a large-scale and structurally diverse benchmark \cite{freitas2021largescaledatabasegraphrepresentation}. \textbf{REDDIT-5K} and \textbf{REDDIT-MULTI-12K} model Reddit discussion threads as user-interaction graphs and require predicting the originating subreddit \cite{morris2020_tudatasets}. \textbf{COLLAB} contains collaboration ego-networks of scientists, with the goal of classifying the research domain from graph topology \cite{morris2020_tudatasets}. \textbf{NCI1} is a molecular benchmark in which the task is to predict anti-cancer activity from atom--bond graphs \cite{morris2020_tudatasets,wale2008_nci_original}. \textbf{ZINC} is a popular molecular regression benchmark based on a 12K subset of the ZINC molecular graph dataset, where the task is to predict constrained solubility from atom--bond graphs \cite{gomez2018_zinc, dwivedi2023_benchmarking}. Finally, \textbf{CO-ER} and \textbf{CO-RB} are graph-level regression benchmarks from GraphBench \cite{stoll2026_graphbenchnextgenerationgraphlearning}, where the task is to predict the maximum independent set objective on Erdős–Rényi random graphs \cite{erdos_1959random} and clique-structured RB graphs \cite{XU_rbgraphs}, respectively.

\section{Experimental Details}
\label{app:experiments}

We conduct all experiments with PyTorch Geometric~\cite{Fey/Lenssen/2019_torch_geometric}, relying on NetworkX~\cite{hagberg2007exploring} and graph-tool~\cite{peixoto_graph-tool_2014} for graph-specific algorithms. Computations are conducted on a single NVIDIA A100 GPU with 80GB VRAM~\footnote{A single model training run takes 10 to 30 minutes on synthetic and around 20 to 70 minutes on real-world datasets.}. 

\paragraph{Model Architecture.}
We combine all of our processing layers with a standardized base model architecture. Initial node features undergo a linear projection to the embedding space. Subgraph node markings (if used) are also projected and summed with the node embeddings. These features, defined either in node space or in subgraph-node space, are then transformed sequentially by the corresponding layer using a GIN with 2-layer MLPs. After each convolution, we apply batch normalization \cite{ioffe2015_batchnorm}, ReLU activation, dropout, and a residual connection. We use a 2-layer MLP to transform aggregated subgraph features as in Eq.~\eqref{eq:readout} and a 3-layer MLP to produce the final graph-level predictions, with all MLPs containing batch normalization, ReLU, and dropout. Unless otherwise indicated, we use mean aggregation throughout for stability, including for graph-level and supernode-level readout as well as for subgraph feature aggregation in Eq.~\eqref{eq:dss_gnn} in \psgnn.

\paragraph{Experimental Setup.}
All models are matched in depth and overall parameter count, where appropriate. Generally, our goal is not to maximize absolute performance, but to compare the three model architectures in a controlled setting. For all deterministic pooling models, we use a single pooling step, ensuring a like-for-like comparison with \psgnn\ and allowing subgraph assignments to be precomputed. As discussed in Section~\ref{sec:experiments}, we compare all models under the same base policy $\pi$ to isolate the effect of architecture, with \psgnn\ using the padded policy $\hat{\pi}$. Hyperparameters are selected based on validation performance. Across different seeds, we use a random 80\%-10\%-10\% train-validation-test split, or a fixed split when one is provided.

\paragraph{Implementation of Subgraph Policies.} We do not tune subgraph-policy hyperparameters and instead use a small set of fixed implementation defaults. Leiden is run with its default resolution parameter of 1.0. For CurvPool, we threshold Balanced Forman Curvature at the 0.8 quantile (which we compared against mean-thresholding), retain edges above this threshold, and merge overlapping selected edges into core clusters; uncovered nodes are subsequently added through a fallback coverage step. For CliquePool, we compute maximal cliques, then resolve overlaps by keeping each node only in its largest incident clique(s), iterating this pruning until stable. In practice, we add node markings beyond the core/halo indicators whenever they are naturally available from the extraction procedure. 

\subsection{BREC}
\label{app:experiments_brec}

We largely follow the original BREC protocol \cite{wang2023_brec}. For each pair of graphs, we train a model by optimizing the cosine embedding loss for a maximum of 40 epochs using Adam \cite{kingma2014adam} with learning rate $10^{-4}$ and weight decay $10^{-4}$. We also replace all batch normalization layers with GraphNorm layers \cite{cai2021_graphnorm} due to numerical stability issues. Finally, we use a sum-based graph-level readout.

After some heuristic tuning, we settled on 8 layers throughout. Table~\ref{tab:layer_allocations} shows how these layers are allocated for the \poolgnn\ and \ipgnn\ models. With this setup, our \poolgnn\ baseline already achieves stronger results than previous empirical investigations into expressive graph pooling (see Table 1 in \cite{lachi2025_expressive_pooling}), supporting our hyperparameter choices and the use of a single pooling step. Intuitively, if a first pooling step does not already improve separation on BREC, additional pooling steps are unlikely to provide additional help.

\begin{table*}[ht]
\centering
\caption{\textbf{Layer allocations across datasets and model families.} For \ipgnn\ and \poolgnn, layers are split into \emph{pre}, \emph{within}, and \emph{post}; for \textbf{Hierarchical}, they are split into \emph{pre}, \emph{between}, and \emph{post}. \psgnn\ and any MPNN baselines always use the full stack, i.e., total depth. \emph{between} layers are multiplied by the number of pooling steps.}
\label{tab:layer_allocations}
\setlength{\tabcolsep}{0pt}
\renewcommand{\arraystretch}{1.10}
\resizebox{\linewidth}{!}{%
\begin{tabular}{@{}
l
@{\hspace{1.2em}}
>{\centering\arraybackslash}m{1.35cm}
@{\hspace{1.5em}}
>{\centering\arraybackslash}m{0.82cm}
>{\centering\arraybackslash}m{0.82cm}
>{\centering\arraybackslash}m{0.82cm}
@{\hspace{1.5em}}
>{\centering\arraybackslash}m{0.82cm}
>{\centering\arraybackslash}m{0.82cm}
>{\centering\arraybackslash}m{0.82cm}
@{\hspace{1.5em}}
>{\centering\arraybackslash}m{0.82cm}
>{\centering\arraybackslash}m{0.82cm}
>{\centering\arraybackslash}m{0.82cm}
@{}}
\toprule
&
\textbf{Total Depth} &
\multicolumn{3}{c}{\textbf{\ipgnn}} &
\multicolumn{3}{c}{\textbf{\poolgnn}} &
\multicolumn{3}{c}{\textbf{Hierarchical}} \\
\cmidrule(l{2pt}r{2pt}){3-5}
\cmidrule(l{2pt}r{2pt}){6-8}
\cmidrule(l{2pt}r{2pt}){9-11}
\textbf{Dataset} & &
\multicolumn{1}{c}{\emph{pre}} &
\multicolumn{1}{c}{\emph{within}} &
\multicolumn{1}{c}{\emph{post}} &
\multicolumn{1}{c}{\emph{pre}} &
\multicolumn{1}{c}{\emph{within}} &
\multicolumn{1}{c}{\emph{post}} &
\multicolumn{1}{c}{\emph{pre}} &
\multicolumn{1}{c}{\emph{between}} &
\multicolumn{1}{c}{\emph{post}} \\
\midrule[0.75pt]

\multicolumn{11}{@{}l}{\textit{Hierarchical Structure}} \\
COLLAB  & \textbf{6} & 2 & 2 & 2 & 3 & 0 & 3 & 2 & 2x2 & 0 \\
REDDIT-12K  & \textbf{6} & 2 & 2 & 2 & 3 & 0 & 3 & 2 & 2x2 & 0 \\
REDDIT-5K  & \textbf{6} & 2 & 2 & 2 & 3 & 0 & 3 & 2 & 2x2 & 0 \\
MalNet  & \textbf{6} & 2 & 2 & 2 & 3 & 0 & 3 & 2 & 2x2 & 0 \\
CO-RB      & \textbf{6} & 2 & 2 & 2 & 3 & 0 & 3 & -- & -- & -- \\
\midrule

\multicolumn{11}{@{}l}{\textit{Flat Structure}} \\
NCI1    & \textbf{6} & 2 & 2 & 2 & 3 & 0 & 3 & 2 & 2x2 & 0 \\
ZINC    & \textbf{8} & 4 & 2 & 2 & 5 & 0 & 3 & 4 & 2x2 & 0 \\
CO-ER & \textbf{6} & 2 & 2 & 2 & 3 & 0 & 3 & -- & -- & -- \\
\midrule

\multicolumn{11}{@{}l}{\textit{Synthetic, structure-controlled datasets}} \\
BREC                 & \textbf{8} & 2 & 3 & 3 & 4 & 0 & 4 & -- & -- & -- \\
Macro-Hierarchies    & \textbf{8} & 4 & 2 & 2 & 5 & 0 & 3 & 4 & 2x2 & 2 \\
Granular-Hierarchies & \textbf{8} & 4 & 2 & 2 & 5 & 0 & 3 & 4 & 2x2 & 2 \\
\bottomrule
\end{tabular}%
}
\end{table*}

\subsection{Synthetic Datasets}
\label{app:experiment_synthetic}

We performed a small hyperparameter sweep over the number of layers $\{8, 12\}$ and embedding size $\{64, 128\}$, evaluating each setting on two seeds. We find that larger models achieve slightly better absolute results, but the relative ordering of the models and the performance gaps between them remain largely unchanged (see Figure~\ref{fig:synthetic_results}).
We therefore settle on 8 layers with an embedding size of 64. To roughly match the number of parameters, we reduce the embedding size of \psgnn\ to 48. Exact layer allocations for each model are presented in Table~\ref{tab:layer_allocations}. We use AdamW \cite{loshchilov2017_adamw} for optimization. All other optimization and model hyperparameters are detailed in Table~\ref{tab:dataset_hyperparameters}.

The GPS~\cite{rampavsek2022_GPS} baseline uses a GIN layer as the local message-passing component and employs Performer attention~\cite{choromanski2020_performers} for scalability.
We set the embedding size to 18 to closely match the parameter budget of the other baselines.
We further add Random Walk Positional/Structural Encodings~\cite{dwivedi2021_rwpe} with length 16, a standard structural encoding for graph transformers. In preliminary experiments, GPS without positional/structural encodings performed substantially worse on our synthetic tasks.

\begin{table*}[ht]
\centering
\caption{\textbf{Dataset-specific training and architecture hyperparameters.} Key hyperparameters used for each benchmark dataset. For BREC, no patience-based early stopping is used; instead, training stops once the loss falls below 0.2 or after 40 epochs.}
\label{tab:dataset_hyperparameters}
\setlength{\tabcolsep}{5pt}
\resizebox{\textwidth}{!}{%
\begin{tabular}{@{}
l
S[table-format=2.0]
S[table-format=3.0]
S[table-format=3.0]
c
c
S[table-format=1.2]
S[table-format=1.2]
c
c
@{}}
\toprule
\textbf{Dataset} &
\shortstack[c]{\textbf{\#}\\\textbf{Layers}} &
\shortstack[c]{\textbf{Max}\\\textbf{Epochs}} &
\shortstack[c]{\textbf{Batch}\\\textbf{Size}} &
\shortstack[c]{\textbf{Learning}\\\textbf{Rate}} &
\shortstack[c]{\textbf{Weight}\\\textbf{Decay}} &
\shortstack[c]{\textbf{Layers}\\\textbf{Dropout}} &
\shortstack[c]{\textbf{Head}\\\textbf{Dropout}} &
\shortstack[c]{\textbf{Early stop}\\\textbf{Patience}} &
\shortstack[c]{\textbf{\#}\\\textbf{Seeds}} \\
\midrule[0.75pt]

\multicolumn{9}{@{}l}{\textit{Hierarchical Structure}} \\
COLLAB           & 6  & 300 & 32  & $10^{-3}$          & $0$       & 0.05 & 0.10 & 50 & 5 \\
REDDIT-12K           & 6  & 300 & 16  & $10^{-3}$          & $0$       & 0.05 & 0.10 & 50 & 5 \\
REDDIT-5K        & 6 & 300 & 16 & $10^{-3}$ & $0$       & 0.05 & 0.10 & 50 & 5 \\
MalNet           & 6  & 300 & 16  & $10^{-3}$          & $0$       & 0.05 & 0.10 & 50 & 5 \\
CO-RB            & 6 & 300 & 32 & $10^{-3}$ & $10^{-5}$ & 0.00 & 0.10 & 50 & 3 \\
\midrule

\multicolumn{9}{@{}l}{\textit{Flat Structure}} \\
NCI1             & 6  & 300 & 32  & $10^{-3}$          & $0$       & 0.00 & 0.10 & 50 & 10 \\
ZINC             & 8 & 500 & 64  & $10^{-3}$          & $10^{-5}$ & 0.00 & 0.00 & 100 & 3 \\
CO-ER            & 6 & 300 & 32 & $10^{-3}$ & $10^{-5}$ & 0.00 & 0.10 & 50 & 3 \\
\midrule

\multicolumn{9}{@{}l}{\textit{Synthetic, structure-controlled datasets}} \\
BREC             & 8 & 40  & 16  & $10^{-4}$          & $10^{-4}$ & 0.00 & 0.00 & --  & 3 \\
Macro-Hierarchies      & 8  & 250 & 128 & $5 \times 10^{-4}$ & $10^{-5}$ & 0.00 & 0.00 & 50 & 5 \\
Granular-Hierarchies & 8 & 250 & 128 & $5 \times 10^{-4}$ & $10^{-5}$ & 0.00 & 0.00 & 50 & 3 \\
\bottomrule
\end{tabular}%
}
\end{table*}

We also tuned the number of pooling steps in $\{1, 2\}$ and the pooling strength using either ratios $\{0.1, 0.5\}$ or a fixed target size of 24 nodes, which roughly corresponds to the number of micro-communities on the \textsc{Macro-Hierarchies} dataset. We find that a pooling ratio of $0.5$ consistently performs best. For all methods except TopKPool, a single pooling step is optimal.

\subsection{Real-World Datasets}
\label{app:experiment_real}

For the real-world benchmarks, we use 6 layers on all datasets except ZINC, where 8 layers is more appropriate. Model-specific allocations of layers are reported in Table~\ref{tab:layer_allocations}. All other hyperparameters used are detailed in Table~\ref{tab:dataset_hyperparameters}. Given this model specification, we set the embedding dimension such that each model fits within a 500K parameter budget. As ZINC uses edge features, we first embed edge features and subsequently employ the GINE layer \cite{hu2020strategies_GINE} for processing. For datasets without node features, we use constant scalars.

We tune the number of pooling steps in $\{1,2\}$ for all hierarchical GNNs and find that using two pooling steps is generally beneficial. We further ablate the use of subgraph node markings and the edge-covering policy $\hat{\pi}$. 
Table~\ref{tab:padding_marking_ablation} presents the result of this experiment. The base model always includes node markings and the padded policy. The impact on performance depends on the dataset and subgraph-policy combination. In most cases, however, the effect is small and remains within one standard deviation of the base model, with the clearest exception being clique-based policies on COLLAB.

\begin{table}[ht]
\centering
\caption{\textbf{Padding and node-marking impact for \psgnn\ on real-world datasets.} The base row is the mean-aggregation configuration with node markings enabled and default padding. Italic rows report the change relative to the base. For accuracy metrics, positive values indicate improvement; for ZINC, negative values indicate improvement because lower MAE is better. Models without padding also contain no node markings, as there is no core vs. halo distinction. Empty cells indicate that a configuration was not tested due to computational limitations.}
\label{tab:padding_marking_ablation}
\setlength{\tabcolsep}{5pt}
\resizebox{\linewidth}{!}{%
\begin{tabular}{@{} l l c c c c c @{}}
\toprule
& & \multicolumn{3}{c}{\textbf{Hierarchical Structure}} & \multicolumn{2}{c}{\textbf{Flat Structure}} \\
\cmidrule(lr){3-5}\cmidrule(lr){6-7}
\textbf{Model} & \textbf{Subgraph Policy} &
\shortstack[c]{\textbf{COLLAB}\\ \textit{\footnotesize Acc. [\%] ($\uparrow$)}} &
\shortstack[c]{\textbf{REDDIT-12K}\\ \textit{\footnotesize Acc. [\%] ($\uparrow$)}} &
\shortstack[c]{\textbf{MalNet}\\ \textit{\footnotesize Acc. [\%] ($\uparrow$)}} &
\shortstack[c]{\textbf{NCI1}\\ \textit{\footnotesize Acc. [\%] ($\uparrow$)}} &
\shortstack[c]{\textbf{ZINC}\\ \textit{\footnotesize MAE ($\downarrow$)}} \\
\midrule[0.75pt]
\textbf{\psgnn} & CliquePool & 78.1 $\pm$ 1.3 & 47.0 $\pm$ 3.8 & 90.6 $\pm$ 2.4 & 81.5 $\pm$ 2.0 & 0.181 $\pm$ 0.005 \\
& \textit{- padding} & -4.4 & +0.8 & +1.0 & -0.6 & -0.002 \\
& \textit{- node markings} & -2.3 & +0.8 & +1.0 & +0.0 & +0.010 \\
& CurvPool & 76.8 $\pm$ 1.5 & 49.3 $\pm$ 1.7 & 92.0 $\pm$ 1.9 & 81.2 $\pm$ 1.6 & 0.175 $\pm$ 0.005 \\
& \textit{- padding} & -0.5 & -0.3 & -1.2 & -0.8 & +0.003 \\
& \textit{- node markings} & +0.5 & +0.8 & -0.7 & -0.8 & -0.001 \\
& Leiden & 76.4 $\pm$ 2.2 & 50.0 $\pm$ 1.3 & 90.9 $\pm$ 1.8 & 77.8 $\pm$ 2.3 & 0.215 $\pm$ 0.012 \\
& \textit{- padding} & -1.4 & -0.8 & +0.2 & +1.0 & -0.006 \\
& \textit{- node markings} & -1.8 & -0.0 & -0.4 & +2.1 & +0.002 \\
\bottomrule
\end{tabular}%
}
\end{table}

Finally, Table~\ref{tab:training_runtime} reports the average training time per epoch of the real-world datasets NCI1 and REDDIT-12K discussed in Section~\ref{sec:scalability}. Although the subgraph-based architectures introduce additional computational overhead relative to the MPNN baseline, their runtime remains within the same order of magnitude on both datasets, consistent with the theoretical complexity analysis.

\begin{table}[ht]
\centering
\caption{\textbf{Training time per epoch on NCI1 and REDDIT-12K.}
Mean training time in seconds per epoch ($\pm$ standard deviation)
for the datasets considered in Figure~7.}
\label{tab:training_runtime}
\setlength{\tabcolsep}{8pt}
\begin{tabular}{@{} l l c c @{}}
\toprule
\textbf{Model} &
\textbf{Subgraph Policy} &
\textbf{NCI1} &
\textbf{REDDIT-12K} \\
\midrule

\textbf{\psgnn}
& CliquePool & 2.47 $\pm$ 0.12 & 16.11 $\pm$ 0.89 \\
& CurvPool   & 2.75 $\pm$ 0.18 & 18.95 $\pm$ 2.88 \\
& Leiden     & 2.45 $\pm$ 0.09 & 17.11 $\pm$ 0.75 \\

\midrule

\textbf{\ipgnn}
& CliquePool & 2.15 $\pm$ 0.17 & 15.04 $\pm$ 2.62 \\
& CurvPool   & 2.08 $\pm$ 0.10 & 15.05 $\pm$ 1.30 \\
& Leiden     & 2.04 $\pm$ 0.04 & 13.79 $\pm$ 1.49 \\

\midrule

\textbf{\poolgnn}
& CliquePool & 2.08 $\pm$ 0.15 & 13.41 $\pm$ 1.92 \\
& CurvPool   & 2.02 $\pm$ 0.12 & 14.68 $\pm$ 2.37 \\
& Leiden     & 1.99 $\pm$ 0.05 & 14.06 $\pm$ 0.96 \\

\midrule

\textbf{MPNN (GIN)}
& -- & 1.16 $\pm$ 0.05 & 9.16 $\pm$ 0.54 \\

\textbf{Egonet}
& -- & 2.52 $\pm$ 0.16 & -- \\

\bottomrule
\end{tabular}
\end{table}

\section{Additional Results for Other Policies}
\label{app:additional_results}

There are not many widely used structure-based pooling schemes in the graph pooling literature \cite{lachi2025_expressive_pooling}. To expand our evaluation, we therefore repeat our experiments with three additional policies described below. 

\begin{itemize}
    \item \textbf{CliquePool-Maximal} is our overlapping variant of CliquePool, in which we keep all maximal cliques as subgraphs instead of resolving overlaps. In the padded version, we add any missing edge to every incident clique-subgraph rather than performing a full 1-hop expansion, so the resulting subgraphs are no longer uniform and contain additional structural variation.
    \item \textbf{ACL} applies the Andersen--Chung--Lang localized clustering algorithm \cite{andersen2006_acl} iteratively, each time seeding it at the highest-degree node that has not yet been assigned to a cluster. In detail, clusters are extracted from a personalized PageRank sweep and constrained to cover between 0.02 and 0.3 of the graph volume; if no sweep prefix falls in this range, we choose the prefix closest to the target range. For computational reasons, we cap the iterative ACL procedure at 1000 extracted subgraphs, after which any remaining uncovered nodes are assigned through a rescue step.
    \item \textbf{SeededHops} is our iterative seed-based variant, which forms a 1-hop egonet around the highest-degree node that has not yet been assigned to a cluster, and repeats this process until all nodes are assigned.
\end{itemize}

Note that the latter two are seed-based policies, where we seed a subgraph extraction logic at some node and repeat until coverage is achieved. These are closer in spirit to the egonet baseline, which extracts $2$-hop expansions around \emph{every} node. 

These policies also give the opportunity to benchmark the usage of additional subgraph node features. In particular, for seed-based policies we add a root-node indicator together with distance-to-root features. For ACL, we additionally include the normalized personalized PageRank score of each node within the extracted cluster. 

\begin{figure*}[ht]
    \centering
    \includegraphics[width=1.\linewidth]{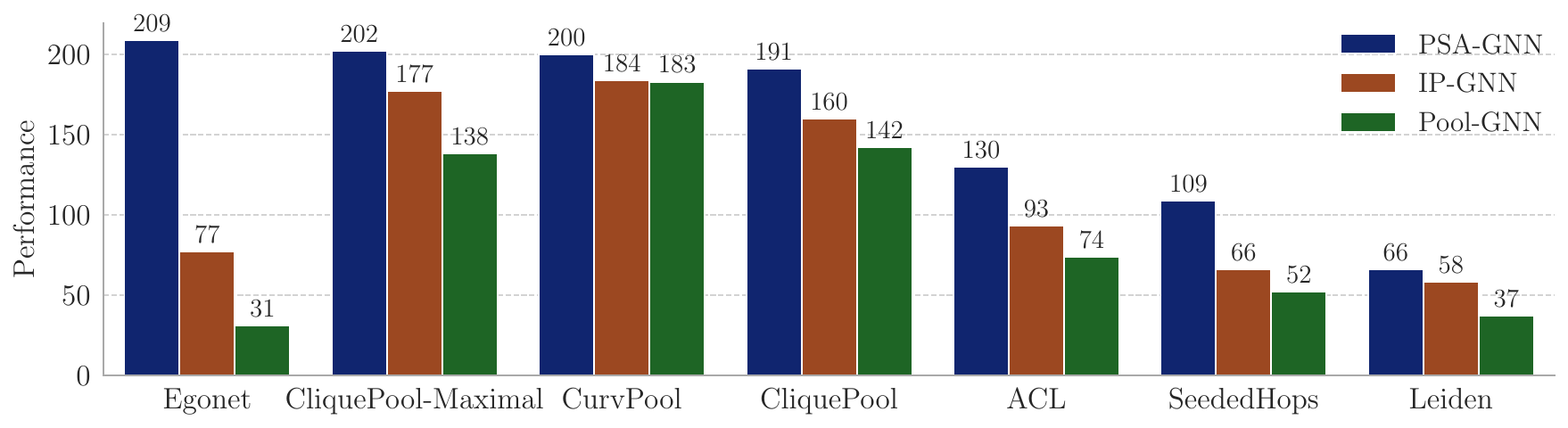}
    \caption{\textbf{Number of BREC graph pairs (out of 400) distinguished by each architecture-policy combination.} We report the best score achieved over three seeds. }
    \label{fig:brec_appendix_results}
\end{figure*}

Figure~\ref{fig:brec_appendix_results} reports the results for the BREC dataset. These new policies follow the exact same trend as the structure-based policies, with \psgnn\ outperforming \poolgnn\ and \ipgnn\ occupying the middle ground. CliquePool-Maximal stands out as a particularly successful approach. This could be potentially explained by the increase in overlap across clusters among all maximal cliques. Notably, this more powerful approach only outperforms standard CliquePool when combined with \psgnn\, but not when processed by \poolgnn, indicating that more powerful policies need more powerful processing approaches.

\begin{figure}[ht]
  \centering
\begin{minipage}[t]{0.03\linewidth}
  \centering
  \rotatebox{90}{\small\textbf{\quad\quad\quad  \textsc{Macro-Hier.}}}
\end{minipage}\hfill
\begin{minipage}[t]{0.95\linewidth}
\includegraphics[width=1.0\linewidth]{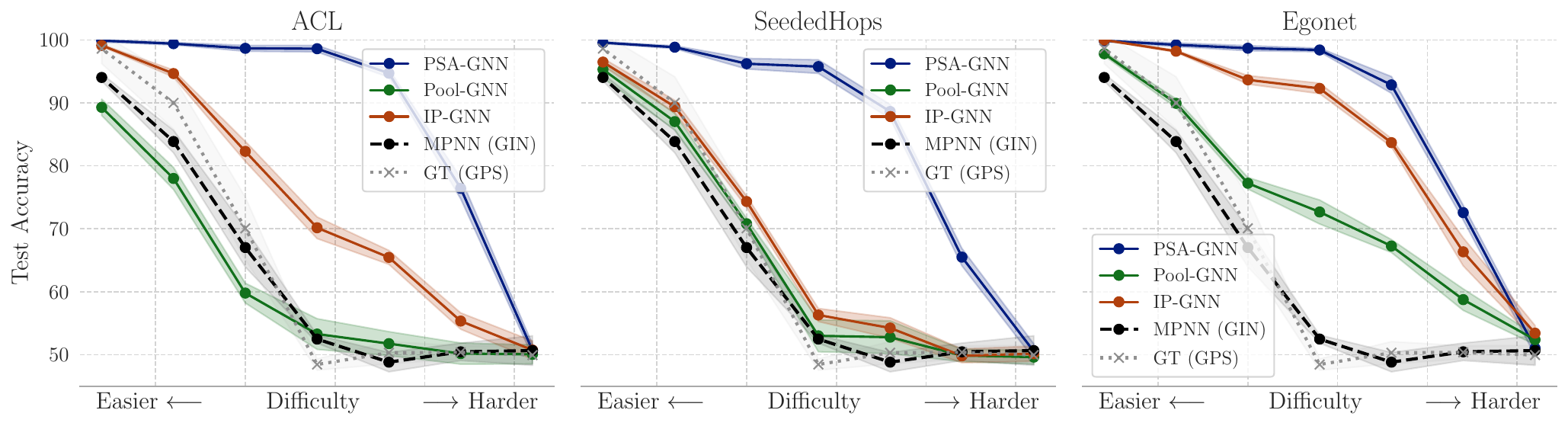}
  \end{minipage}

\begin{minipage}[t]{0.03\linewidth}
  \centering
  \rotatebox{90}{\small\textbf{\quad\quad  \textsc{Granular-Hier.}}}
\end{minipage}\hfill
\begin{minipage}[t]{0.95\linewidth}
\includegraphics[width=1.0\linewidth]{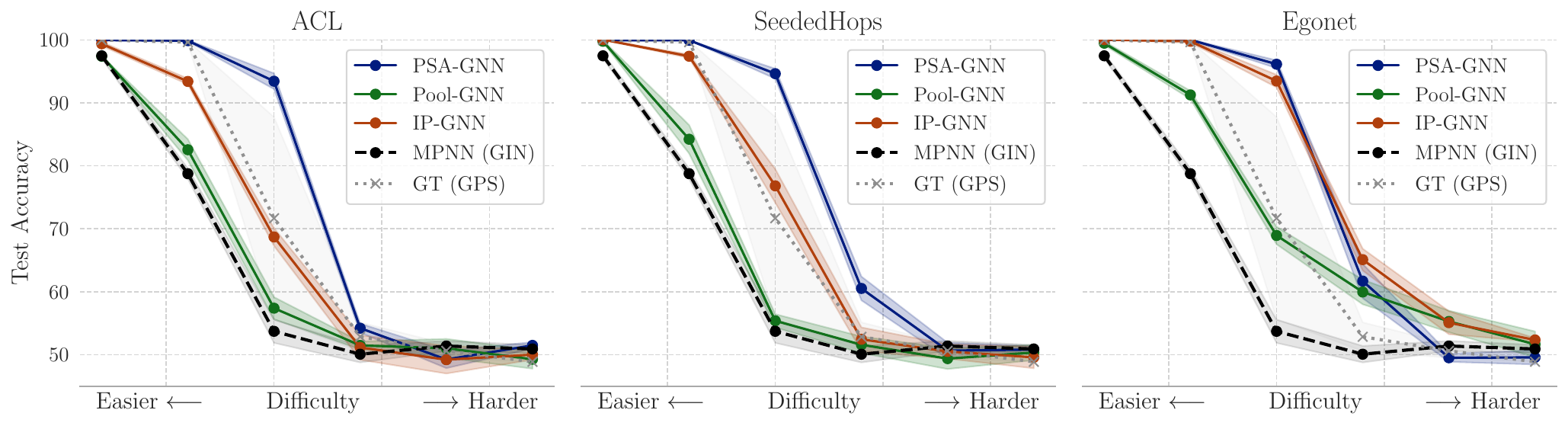}
  \end{minipage}
  
\caption{\textbf{Results on the synthetic hierarchical benchmarks for additional subgraph policies.} We compare immediate coarsening (\poolgnn), the within-subgraph baseline (\ipgnn), and the full pooled-subgraph model (\psgnn) across three additional subgraph policies. }
  \label{fig:appendix_synthetic_results}
\end{figure}

Next, we repeated the experiment on our synthetic datasets, with results presented in Figure~\ref{fig:appendix_synthetic_results}, excluding CliquePool-Maximal for computational reasons. Again, the pattern discussed in the main text is clearly confirmed on these additional policies. Note that due to the structure of the dataset, even Egonets arguably group nodes belonging to the same community, as clusters are more densely connected.  

Finally, we present an expanded Table~\ref{tab:appendix_real_world_results} reporting results on real-world benchmarks including our new policies. \psgnn\ remains the top performing model. 
CliquePool-Maximal emerges as a powerful policy, reaching the best result on all hierarchical datasets tested, while remaining computationally feasible.  

\begin{table}[ht]
\centering
\caption{\textbf{Results on real-world benchmark datasets.} The best overall result per dataset is shown in \bestresult{\textbf{brown}}, second best in \secondresult{\textbf{purple}}. Among \psgnn, \ipgnn, and \poolgnn, the best method for a given subgraph policy $\pi$ is shown in \textbf{bold}. The \emph{Avg. Rank} column reports the average rank of these three architecture levels across all available policy--dataset comparisons.}
\label{tab:appendix_real_world_results}
\setlength{\tabcolsep}{5pt}
\resizebox{\linewidth}{!}{%
\begin{tabular}{@{} l l c c c c c c @{}}
\toprule
& & & \multicolumn{3}{c}{\textbf{Hierarchical Structure}} & \multicolumn{2}{c}{\textbf{Flat Structure}} \\
\cmidrule(lr){4-6}\cmidrule(lr){7-8}
\textbf{Model} & \textbf{Subgraph Policy} &
\shortstack[c]{\textbf{Avg.}\\ \textbf{Rank} ($\downarrow$)} &
\shortstack[c]{\textbf{COLLAB}\\ \textit{\footnotesize Acc. [\%] ($\uparrow$)}} &
\shortstack[c]{\textbf{REDDIT-12K}\\ \textit{\footnotesize Acc. [\%] ($\uparrow$)}} &
\shortstack[c]{\textbf{MalNet-T}\\ \textit{\footnotesize Acc. [\%] ($\uparrow$)}} &
\shortstack[c]{\textbf{NCI1}\\ \textit{\footnotesize Acc. [\%] ($\uparrow$)}} &
\shortstack[c]{\textbf{ZINC}\\ \textit{\footnotesize MAE ($\downarrow$)}} \\
\midrule[0.75pt]
\textbf{\psgnn} & CliquePool & \multirow{6}{*}{\textbf{1.34}} & \secondresult{\textbf{78.1 $\pm$ 1.3}} & 47.8 $\pm$ 1.4 & 90.6 $\pm$ 2.4 & \secondresult{\textbf{81.5 $\pm$ 2.5}} & \textbf{0.182 $\pm$ 0.008} \\
 & CliquePool-max &  & \bestresult{\textbf{78.9 $\pm$ 0.7}} & \bestresult{\textbf{51.2 $\pm$ 0.0}} & \bestresult{\textbf{92.3 $\pm$ 1.6}} & 80.9 $\pm$ 1.6 & \textbf{0.179 $\pm$ 0.003} \\
 & CurvPool &  & 76.8 $\pm$ 1.5 & \textbf{50.1 $\pm$ 1.7} & \textbf{92.0 $\pm$ 1.9} & 81.2 $\pm$ 1.6 & \textbf{0.177 $\pm$ 0.005} \\
 & Leiden &  & \textbf{76.4 $\pm$ 2.2} & \textbf{50.0 $\pm$ 1.3} & \textbf{91.1 $\pm$ 1.4} & 80.0 $\pm$ 2.5 & 0.213 $\pm$ 0.007 \\
 & ACL &  & -- & -- & \secondresult{\textbf{92.1 $\pm$ 0.9}} & \textbf{80.9 $\pm$ 2.8} & \textbf{0.189 $\pm$ 0.002} \\
 & SeededHops &  & 72.2 $\pm$ 4.6 & \secondresult{\textbf{50.4 $\pm$ 1.4}} & \textbf{91.4 $\pm$ 2.0} & 81.1 $\pm$ 2.6 & \textbf{0.184 $\pm$ 0.006} \\
\midrule[0.75pt]
\textbf{\ipgnn} & CliquePool & \multirow{6}{*}{2.32} & 75.0 $\pm$ 2.3 & 45.1 $\pm$ 2.1 & \textbf{91.7 $\pm$ 1.7} & 80.6 $\pm$ 2.1 & 0.183 $\pm$ 0.004 \\
 & CliquePool-max &  & 78.1 $\pm$ 0.4 & 49.3 $\pm$ 0.7 & 91.6 $\pm$ 1.4 & 80.2 $\pm$ 2.4 & 0.187 $\pm$ 0.002 \\
 & CurvPool &  & \textbf{77.4 $\pm$ 0.9} & 42.0 $\pm$ 2.5 & 89.3 $\pm$ 2.0 & 80.3 $\pm$ 1.9 & 0.184 $\pm$ 0.005 \\
 & Leiden &  & 73.6 $\pm$ 3.1 & 45.6 $\pm$ 0.8 & 90.7 $\pm$ 1.1 & \textbf{80.1 $\pm$ 2.6} & \textbf{0.186 $\pm$ 0.004} \\
 & ACL &  & -- & -- & 90.5 $\pm$ 1.5 & 80.7 $\pm$ 3.6 & 0.248 $\pm$ 0.005 \\
 & SeededHops &  & 71.5 $\pm$ 5.9 & 48.8 $\pm$ 1.0 & 89.6 $\pm$ 2.3 & 81.5 $\pm$ 1.6 & 0.209 $\pm$ 0.002 \\
\midrule[0.75pt]
\textbf{\poolgnn} & CliquePool & \multirow{6}{*}{2.34} & 70.4 $\pm$ 4.1 & \textbf{48.3 $\pm$ 1.7} & 90.2 $\pm$ 1.6 & 81.0 $\pm$ 2.4 & \textbf{0.182 $\pm$ 0.004} \\
 & CliquePool-max &  & 76.0 $\pm$ 2.3 & 46.1 $\pm$ 6.4 & 90.2 $\pm$ 2.3 & 80.3 $\pm$ 2.4 & 0.181 $\pm$ 0.004 \\
 & CurvPool &  & 71.1 $\pm$ 4.5 & 47.4 $\pm$ 0.4 & 91.3 $\pm$ 1.1 & \textbf{81.3 $\pm$ 2.0} & 0.178 $\pm$ 0.004 \\
 & Leiden &  & 74.1 $\pm$ 2.3 & 44.7 $\pm$ 1.6 & 87.9 $\pm$ 1.5 & 79.9 $\pm$ 2.0 & 0.257 $\pm$ 0.013 \\
 & ACL &  & -- & -- & 89.8 $\pm$ 2.0 & 80.8 $\pm$ 2.1 & 0.270 $\pm$ 0.005 \\
 & SeededHops &  & \textbf{73.2 $\pm$ 2.9} & 48.4 $\pm$ 1.9 & 90.4 $\pm$ 1.6 & \bestresult{\textbf{81.6 $\pm$ 2.0}} & 0.229 $\pm$ 0.003 \\
\midrule[0.75pt]
\deemph{\textbf{Hierarchical GNN}} & \deemph{ASAPool} & \deemph{--} & \deemph{71.6 $\pm$ 5.4} & \deemph{49.5 $\pm$ 0.2} & \deemph{88.2 $\pm$ 3.1} & \deemph{81.2 $\pm$ 2.0} & \deemph{--} \\
\deemph{} & \deemph{SAGPool} & \deemph{--} & \deemph{70.6 $\pm$ 4.2} & \deemph{46.6 $\pm$ 1.8} & \deemph{90.1 $\pm$ 1.1} & \deemph{80.1 $\pm$ 1.9} & \deemph{0.416 $\pm$ 0.107} \\
\deemph{} & \deemph{MinCutPool} & \deemph{--} & \deemph{--} & \deemph{--} & \deemph{--} & \deemph{80.2 $\pm$ 2.0} & \deemph{\secondresult{0.161 $\pm$ 0.003}} \\
\deemph{} & \deemph{DiffPool} & \deemph{--} & \deemph{--} & \deemph{--} & \deemph{--} & \deemph{81.0 $\pm$ 1.7} & \deemph{\bestresult{0.147 $\pm$ 0.005}} \\
\midrule
\deemph{\textbf{Baselines}} & \deemph{Egonet} & \deemph{--} & \deemph{--} & \deemph{--} & \deemph{91.4 $\pm$ 1.4} & \deemph{\bestresult{81.6 $\pm$ 2.4}} & \deemph{0.403 $\pm$ 0.060} \\
\deemph{} & \deemph{MPNN (GIN)} & \deemph{--} & \deemph{74.9 $\pm$ 3.7} & \deemph{48.3 $\pm$ 2.3} & \deemph{91.0 $\pm$ 1.6} & \deemph{80.9 $\pm$ 2.5} & \deemph{0.172 $\pm$ 0.005} \\
\bottomrule
\end{tabular}%
}
\end{table}

\clearpage

%% file: main.bbl
\begin{thebibliography}{61}
\providecommand{\natexlab}[1]{#1}
\providecommand{\url}[1]{\texttt{#1}}
\expandafter\ifx\csname urlstyle\endcsname\relax
  \providecommand{\doi}[1]{doi: #1}\else
  \providecommand{\doi}{doi: \begingroup \urlstyle{rm}\Url}\fi

\bibitem[Scarselli et~al.(2009)Scarselli, Gori, Tsoi, Hagenbuchner, and Monfardini]{scarselli2008graph}
Franco Scarselli, Marco Gori, Ah~Chung Tsoi, Markus Hagenbuchner, and Gabriele Monfardini.
\newblock The graph neural network model.
\newblock \emph{IEEE Transactions on Neural Networks}, 20\penalty0 (1):\penalty0 61--80, 2009.

\bibitem[Wu et~al.(2021)Wu, Pan, Chen, Long, Zhang, and Yu]{wu2020_comprehensive}
Zonghan Wu, Shirui Pan, Fengwen Chen, Guodong Long, Chengqi Zhang, and Philip~S. Yu.
\newblock A comprehensive survey on graph neural networks.
\newblock \emph{IEEE Transactions on Neural Networks and Learning Systems}, 32\penalty0 (1):\penalty0 4--24, 2021.

\bibitem[Gilmer et~al.(2017)Gilmer, Schoenholz, Riley, Vinyals, and Dahl]{gilmer2017_neural}
Justin Gilmer, Samuel~S Schoenholz, Patrick~F Riley, Oriol Vinyals, and George~E Dahl.
\newblock Neural message passing for quantum chemistry.
\newblock In \emph{International Conference on Machine Learning}, pages 1263--1272, 2017.

\bibitem[Xu et~al.(2019)Xu, Hu, Leskovec, and Jegelka]{xu2018_GIN}
Keyulu Xu, Weihua Hu, Jure Leskovec, and Stefanie Jegelka.
\newblock How powerful are graph neural networks?
\newblock In \emph{International Conference on Learning Representations}, 2019.

\bibitem[Girvan and Newman(2002)]{girvan2002_community_organization}
Michelle Girvan and Mark~EJ Newman.
\newblock Community structure in social and biological networks.
\newblock \emph{Proceedings of the National Academy of Sciences}, 99\penalty0 (12):\penalty0 7821--7826, 2002.

\bibitem[Ravasz and Barab{\'a}si(2003)]{ravasz2003_hierarchical_organization}
Erzs{\'e}bet Ravasz and Albert-L{\'a}szl{\'o} Barab{\'a}si.
\newblock Hierarchical organization in complex networks.
\newblock \emph{Physical Review E}, 67\penalty0 (2):\penalty0 026112, 2003.

\bibitem[Clauset et~al.(2008)Clauset, Moore, and Newman]{clauset2008_hierarchical_helps}
Aaron Clauset, Cristopher Moore, and Mark~EJ Newman.
\newblock Hierarchical structure and the prediction of missing links in networks.
\newblock \emph{Nature}, 453\penalty0 (7191):\penalty0 98--101, 2008.

\bibitem[Liu et~al.(2023)Liu, Zhan, Wu, Li, Du, Hu, Liu, and Tao]{liu__2023_pooling_lit_review}
Chuang Liu, Yibing Zhan, Jia Wu, Chang Li, Bo~Du, Wenbin Hu, Tongliang Liu, and Dacheng Tao.
\newblock Graph pooling for graph neural networks: Progress, challenges, and opportunities.
\newblock In \emph{Proceedings of the Thirty-Second International Joint Conference on Artificial Intelligence}, pages 6712--6722, 2023.

\bibitem[Grattarola et~al.(2022)Grattarola, Zambon, Bianchi, and Alippi]{grattarola_understanding_gnn_pooling_2022}
Daniele Grattarola, Daniele Zambon, Filippo~Maria Bianchi, and Cesare Alippi.
\newblock Understanding {Pooling} in {Graph} {Neural} {Networks}.
\newblock \emph{IEEE Transactions on Neural Networks and Learning Systems}, 2022.

\bibitem[Mesquita et~al.(2020)Mesquita, Souza, and Kaski]{mesquita_rethinking_2020}
Diego Mesquita, Amauri Souza, and Samuel Kaski.
\newblock Rethinking pooling in graph neural networks.
\newblock In \emph{Advances in Neural Information Processing Systems}, volume~33, pages 2220--2231, 2020.

\bibitem[Lachi et~al.(2025)Lachi, Moallemy-Oureh, Roth, and Welke]{lachi2025_expressive_pooling}
Veronica Lachi, Alice Moallemy-Oureh, Andreas Roth, and Pascal Welke.
\newblock Expressive pooling for graph neural networks.
\newblock \emph{Transactions on Machine Learning Research}, 2025.

\bibitem[Bianchi and Lachi(2023)]{bianchi_expressive_2023}
Filippo~Maria Bianchi and Veronica Lachi.
\newblock The expressive power of pooling in graph neural networks.
\newblock In \emph{Advances in Neural Information Processing Systems}, 2023.

\bibitem[Wang et~al.(2024)Wang, Luo, Shen, Zhang, Qin, Heng, and Luo]{wang2024_comprehensive_pooling_benchmark}
Pengyun Wang, Junyu Luo, Yanxin Shen, Ming Zhang, Shaoen Qin, Siyu Heng, and Xiao Luo.
\newblock A comprehensive graph pooling benchmark: Effectiveness, robustness and generalizability.
\newblock \emph{arXiv preprint arXiv:2406.09031}, 2024.

\bibitem[Errica et~al.(2020)Errica, Podda, Bacciu, and Micheli]{errica2022_fair_comparison}
Federico Errica, Marco Podda, Davide Bacciu, and Alessio Micheli.
\newblock A fair comparison of graph neural networks for graph classification.
\newblock In \emph{International Conference on Learning Representations}, 2020.

\bibitem[Bevilacqua et~al.(2022)Bevilacqua, Frasca, Lim, Srinivasan, Cai, Balamurugan, Bronstein, and Maron]{bevilacqua2021_ESAN}
Beatrice Bevilacqua, Fabrizio Frasca, Derek Lim, Balasubramaniam Srinivasan, Chen Cai, Gopinath Balamurugan, Michael~M. Bronstein, and Haggai Maron.
\newblock Equivariant subgraph aggregation networks.
\newblock In \emph{International Conference on Learning Representations}, 2022.

\bibitem[Zhao et~al.(2022)Zhao, Jin, Akoglu, and Shah]{zhao2021_gnnak}
Lingxiao Zhao, Wei Jin, Leman Akoglu, and Neil Shah.
\newblock From stars to subgraphs: Uplifting any {GNN} with local structure awareness.
\newblock In \emph{International Conference on Learning Representations}, 2022.

\bibitem[Zhang and Li(2021)]{zhang2021_nestedGNN}
Muhan Zhang and Pan Li.
\newblock Nested graph neural networks.
\newblock In \emph{Advances in Neural Information Processing Systems}, volume~34, pages 15734--15747, 2021.

\bibitem[Frasca et~al.(2022)Frasca, Bevilacqua, Bronstein, and Maron]{frasca2022_sun}
Fabrizio Frasca, Beatrice Bevilacqua, Michael Bronstein, and Haggai Maron.
\newblock Understanding and extending subgraph {GNNs} by rethinking their symmetries.
\newblock In \emph{Advances in Neural Information Processing Systems}, volume~35, pages 31376--31390, 2022.

\bibitem[Zaheer et~al.(2017)Zaheer, Kottur, Ravanbakhsh, Poczos, Salakhutdinov, and Smola]{zaheer2017_deepsets}
Manzil Zaheer, Satwik Kottur, Siamak Ravanbakhsh, Barnabas Poczos, Russ~R Salakhutdinov, and Alexander~J Smola.
\newblock Deep sets.
\newblock In \emph{Advances in Neural Information Processing Systems}, volume~30, 2017.

\bibitem[Sbicego et~al.(2025)Sbicego, {\"O}{\u g}{\"u}t, Madeira, Qin, Thanou, and Frossard]{sbicego2025role}
Luca Sbicego, Sevda {\"O}{\u g}{\"u}t, Manuel Madeira, Yiming Qin, Dorina Thanou, and Pascal Frossard.
\newblock On the role of structure in hierarchical graph neural networks.
\newblock In \emph{Proceedings on ``I Can't Believe It's Not Better: Challenges in Applied Deep Learning'' at ICLR 2025 Workshops}, volume 296 of \emph{Proceedings of Machine Learning Research}, pages 141--150. PMLR, 2025.

\bibitem[Holland et~al.(1983)Holland, Laskey, and Leinhardt]{holland1983_sbm}
Paul~W Holland, Kathryn~Blackmond Laskey, and Samuel Leinhardt.
\newblock Stochastic blockmodels: First steps.
\newblock \emph{Social Networks}, 5\penalty0 (2):\penalty0 109--137, 1983.

\bibitem[Karrer and Newman(2011)]{karrer2011_degree_sbm}
Brian Karrer and Mark~EJ Newman.
\newblock Stochastic blockmodels and community structure in networks.
\newblock \emph{Physical Review E}, 83\penalty0 (1):\penalty0 016107, 2011.

\bibitem[Peixoto(2014{\natexlab{a}})]{peixoto2014_hierarchicalsbm}
Tiago~P Peixoto.
\newblock Hierarchical block structures and high-resolution model selection in large networks.
\newblock \emph{Physical Review X}, 4\penalty0 (1):\penalty0 011047, 2014{\natexlab{a}}.

\bibitem[G{\'o}mez-Bombarelli et~al.(2018)G{\'o}mez-Bombarelli, Wei, Duvenaud, Hern{\'a}ndez-Lobato, S{\'a}nchez-Lengeling, Sheberla, Aguilera-Iparraguirre, Hirzel, Adams, and Aspuru-Guzik]{gomez2018_zinc}
Rafael G{\'o}mez-Bombarelli, Jennifer~N Wei, David Duvenaud, Jos{\'e}~Miguel Hern{\'a}ndez-Lobato, Benjam{\'\i}n S{\'a}nchez-Lengeling, Dennis Sheberla, Jorge Aguilera-Iparraguirre, Timothy~D Hirzel, Ryan~P Adams, and Al{\'a}n Aspuru-Guzik.
\newblock Automatic chemical design using a data-driven continuous representation of molecules.
\newblock \emph{ACS Central Science}, 4\penalty0 (2):\penalty0 268--276, 2018.

\bibitem[Morris et~al.(2020)Morris, Kriege, Bause, Kersting, Mutzel, and Neumann]{morris2020_tudatasets}
Christopher Morris, Nils~M Kriege, Franka Bause, Kristian Kersting, Petra Mutzel, and Marion Neumann.
\newblock {TUDataset}: A collection of benchmark datasets for learning with graphs.
\newblock \emph{arXiv preprint arXiv:2007.08663}, 2020.

\bibitem[Freitas et~al.(2021)Freitas, Dong, Neil, and Chau]{freitas2021largescaledatabasegraphrepresentation}
Scott Freitas, Yuxiao Dong, Joshua Neil, and Duen~Horng Chau.
\newblock A large-scale database for graph representation learning.
\newblock In \emph{Proceedings of the Neural Information Processing Systems Track on Datasets and Benchmarks}, volume~1, 2021.

\bibitem[Traag et~al.(2019)Traag, Waltman, and Van~Eck]{traag2019_louvain}
Vincent~A Traag, Ludo Waltman, and Nees~Jan Van~Eck.
\newblock From {Louvain} to {Leiden}: guaranteeing well-connected communities.
\newblock \emph{Scientific Reports}, 9\penalty0 (1):\penalty0 5233, 2019.

\bibitem[Sanders et~al.(2023)Sanders, Roth, and Liebig]{sanders2023_curvpool}
Cedric Sanders, Andreas Roth, and Thomas Liebig.
\newblock Curvature-based pooling within graph neural networks.
\newblock In \emph{Joint European Conference on Machine Learning and Knowledge Discovery in Databases}, pages 471--485, 2023.

\bibitem[Luzhnica et~al.(2019)Luzhnica, Day, and Lio]{luzhnica2019_cliquepool}
Enxhell Luzhnica, Ben Day, and Pietro Lio.
\newblock Clique pooling for graph classification.
\newblock \emph{arXiv preprint arXiv:1904.00374}, 2019.

\bibitem[Ramp{\'a}{\v{s}}ek et~al.(2022)Ramp{\'a}{\v{s}}ek, Galkin, Dwivedi, Luu, Wolf, and Beaini]{rampavsek2022_GPS}
Ladislav Ramp{\'a}{\v{s}}ek, Michael Galkin, Vijay~Prakash Dwivedi, Anh~Tuan Luu, Guy Wolf, and Dominique Beaini.
\newblock Recipe for a general, powerful, scalable graph transformer.
\newblock In \emph{Advances in Neural Information Processing Systems}, volume~35, pages 14501--14515, 2022.

\bibitem[Lee et~al.(2019)Lee, Lee, and Kang]{lee2019_sagpool}
Junhyun Lee, Inyeop Lee, and Jaewoo Kang.
\newblock Self-attention graph pooling.
\newblock In \emph{International Conference on Machine Learning}, pages 3734--3743, 2019.

\bibitem[Ranjan et~al.(2020)Ranjan, Sanyal, and Talukdar]{ranjan2020_asap}
Ekagra Ranjan, Soumya Sanyal, and Partha Talukdar.
\newblock {ASAP}: Adaptive structure aware pooling for learning hierarchical graph representations.
\newblock In \emph{Proceedings of the AAAI Conference on Artificial Intelligence}, volume~34, pages 5470--5477, 2020.

\bibitem[Bianchi et~al.(2020)Bianchi, Grattarola, and Alippi]{bianchi_spectral_2020}
Filippo~Maria Bianchi, Daniele Grattarola, and Cesare Alippi.
\newblock Spectral {Clustering} with {Graph} {Neural} {Networks} for {Graph} {Pooling}.
\newblock In \emph{{International} {Conference} on {Machine} {Learning}}, 2020.

\bibitem[Ying et~al.(2018)Ying, You, Morris, Ren, Hamilton, and Leskovec]{ying_hierarchical_2018}
Zhitao Ying, Jiaxuan You, Christopher Morris, Xiang Ren, Will Hamilton, and Jure Leskovec.
\newblock Hierarchical {Graph} {Representation} {Learning} with {Differentiable} {Pooling}.
\newblock In \emph{Advances in {Neural} {Information} {Processing} {Systems}}, 2018.

\bibitem[Deshpande et~al.(2018)Deshpande, Sen, Montanari, and Mossel]{deshpande2018_contextualSBM}
Yash Deshpande, Subhabrata Sen, Andrea Montanari, and Elchanan Mossel.
\newblock Contextual stochastic block models.
\newblock In \emph{Advances in Neural Information Processing Systems}, volume~31, 2018.

\bibitem[Stoll et~al.(2025)Stoll, Qian, Finkelshtein, Parviz, Weber, Frasca, Shavit, Siraudin, Mielke, Anastacio, Müller, Bechler-Speicher, Bronstein, Galkin, Hoos, Niepert, Perozzi, Tönshoff, and Morris]{stoll2026_graphbenchnextgenerationgraphlearning}
Timo Stoll, Chendi Qian, Ben Finkelshtein, Ali Parviz, Darius Weber, Fabrizio Frasca, Hadar Shavit, Antoine Siraudin, Arman Mielke, Marie Anastacio, Erik Müller, Maya Bechler-Speicher, Michael Bronstein, Mikhail Galkin, Holger Hoos, Mathias Niepert, Bryan Perozzi, Jan Tönshoff, and Christopher Morris.
\newblock {GraphBench}: Next-generation graph learning benchmarking.
\newblock \emph{arXiv preprint arXiv:2512.04475}, 2025.

\bibitem[Erd{\H{o}}s and R{\'e}nyi(1959)]{erdos_1959random}
Paul Erd{\H{o}}s and Alfr{\'e}d R{\'e}nyi.
\newblock On random graphs {I}.
\newblock \emph{Publicationes Mathematicae Debrecen}, 6:\penalty0 290--297, 1959.

\bibitem[Xu et~al.(2007)Xu, Boussemart, Hemery, and Lecoutre]{XU_rbgraphs}
Ke~Xu, Frédéric Boussemart, Fred Hemery, and Christophe Lecoutre.
\newblock Random constraint satisfaction: Easy generation of hard (satisfiable) instances.
\newblock \emph{Artificial Intelligence}, 171\penalty0 (8):\penalty0 514--534, 2007.

\bibitem[Xu et~al.(2026)Xu, Bai, Cui, Li, Du, Lyu, Wang, and Hancock]{xu2026_sshpool}
Zhuo Xu, Lu~Bai, Lixin Cui, Ming Li, Hangyuan Du, Ziyu Lyu, Yue Wang, and Edwin~R Hancock.
\newblock {SSHPool}: The separated subgraph-based hierarchical pooling.
\newblock In \emph{Proceedings of the AAAI Conference on Artificial Intelligence}, volume~40, pages 27405--27413, 2026.

\bibitem[He et~al.(2023)He, Hooi, Laurent, Perold, LeCun, and Bresson]{he2023_mlpmixer}
Xiaoxin He, Bryan Hooi, Thomas Laurent, Adam Perold, Yann LeCun, and Xavier Bresson.
\newblock A generalization of {ViT/MLP-Mixer} to graphs.
\newblock In \emph{International Conference on Machine Learning}, pages 12724--12745, 2023.

\bibitem[Feng et~al.(2019)Feng, You, Zhang, Ji, and Gao]{feng2019_hypergraphNN}
Yifan Feng, Haoxuan You, Zizhao Zhang, Rongrong Ji, and Yue Gao.
\newblock Hypergraph neural networks.
\newblock In \emph{Proceedings of the AAAI Conference on Artificial Intelligence}, volume~33, pages 3558--3565, 2019.

\bibitem[Duta and Lio(2025)]{duta2025_sphinx}
Iulia Duta and Pietro Lio.
\newblock Sphinx: Structural prediction using hypergraph inference network.
\newblock In \emph{International Conference on Machine Learning}, pages 14884--14901, 2025.

\bibitem[Kim et~al.(2024)Kim, Lee, Gao, Antelmi, Polato, and Shin]{kim2024_hypergraph_survey}
Sunwoo Kim, Soo~Yong Lee, Yue Gao, Alessia Antelmi, Mirko Polato, and Kijung Shin.
\newblock A survey on hypergraph neural networks: An in-depth and step-by-step guide.
\newblock In \emph{Proceedings of the 30th ACM SIGKDD Conference on Knowledge Discovery and Data Mining}, pages 6534--6544, 2024.

\bibitem[Wang and Zhang(2024)]{wang2023_brec}
Yanbo Wang and Muhan Zhang.
\newblock An empirical study of realized {GNN} expressiveness.
\newblock In \emph{International Conference on Machine Learning}, pages 52134--52155, 2024.

\bibitem[Arvind et~al.(2020)Arvind, Fuhlbr{\"u}ck, K{\"o}bler, and Verbitsky]{arvind2020_weisfeiler_examples}
Vikraman Arvind, Frank Fuhlbr{\"u}ck, Johannes K{\"o}bler, and Oleg Verbitsky.
\newblock On {Weisfeiler-Leman} invariance: Subgraph counts and related graph properties.
\newblock \emph{Journal of Computer and System Sciences}, 113:\penalty0 42--59, 2020.

\bibitem[Gao and Ji(2019)]{gao2019_topk_graph_unets}
Hongyang Gao and Shuiwang Ji.
\newblock Graph {U-Nets}.
\newblock In \emph{International Conference on Machine Learning}, pages 2083--2092, 2019.

\bibitem[Knyazev et~al.(2019)Knyazev, Taylor, and Amer]{knyazev2019_top_k_2_understanding}
Boris Knyazev, Graham~W Taylor, and Mohamed Amer.
\newblock Understanding attention and generalization in graph neural networks.
\newblock In \emph{Advances in Neural Information Processing Systems}, 2019.

\bibitem[Shi et~al.(2024)Shi, Pan, Hu, and Dokmani{\'c}]{shi2024_homophily_modulation}
Cheng Shi, Liming Pan, Hong Hu, and Ivan Dokmani{\'c}.
\newblock Homophily modulates double descent generalization in graph convolution networks.
\newblock \emph{Proceedings of the National Academy of Sciences}, 121\penalty0 (8):\penalty0 e2309504121, 2024.

\bibitem[Wale et~al.(2008)Wale, Watson, and Karypis]{wale2008_nci_original}
Nikil Wale, Ian~A Watson, and George Karypis.
\newblock Comparison of descriptor spaces for chemical compound retrieval and classification.
\newblock \emph{Knowledge and Information Systems}, 14:\penalty0 347--375, 2008.

\bibitem[Dwivedi et~al.(2023)Dwivedi, Joshi, Luu, Laurent, Bengio, and Bresson]{dwivedi2023_benchmarking}
Vijay~Prakash Dwivedi, Chaitanya~K Joshi, Anh~Tuan Luu, Thomas Laurent, Yoshua Bengio, and Xavier Bresson.
\newblock Benchmarking graph neural networks.
\newblock \emph{Journal of Machine Learning Research}, 24\penalty0 (43):\penalty0 1--48, 2023.

\bibitem[Fey and Lenssen(2019)]{Fey/Lenssen/2019_torch_geometric}
Matthias Fey and Jan~E. Lenssen.
\newblock Fast graph representation learning with {PyTorch Geometric}.
\newblock In \emph{ICLR Workshop on Representation Learning on Graphs and Manifolds}, 2019.

\bibitem[Hagberg et~al.(2008)Hagberg, Schult, and Swart]{hagberg2007exploring}
Aric~A. Hagberg, Daniel~A. Schult, and Pieter~J. Swart.
\newblock Exploring network structure, dynamics, and function using {NetworkX}.
\newblock In \emph{Proceedings of the 7th Python in Science Conference}, pages 11--15, 2008.

\bibitem[Peixoto(2014{\natexlab{b}})]{peixoto_graph-tool_2014}
Tiago~P. Peixoto.
\newblock The graph-tool {Python} library.
\newblock \emph{figshare}, 2014{\natexlab{b}}.
\newblock \doi{10.6084/m9.figshare.1164194}.

\bibitem[Ioffe and Szegedy(2015)]{ioffe2015_batchnorm}
Sergey Ioffe and Christian Szegedy.
\newblock Batch normalization: Accelerating deep network training by reducing internal covariate shift.
\newblock In \emph{International Conference on Machine Learning}, pages 448--456, 2015.

\bibitem[Kingma and Ba(2015)]{kingma2014adam}
Diederik~P. Kingma and Jimmy Ba.
\newblock Adam: A method for stochastic optimization.
\newblock In \emph{International Conference on Learning Representations}, 2015.

\bibitem[Cai et~al.(2021)Cai, Luo, Xu, He, Liu, and Wang]{cai2021_graphnorm}
Tianle Cai, Shengjie Luo, Keyulu Xu, Di~He, Tie-yan Liu, and Liwei Wang.
\newblock {GraphNorm}: A principled approach to accelerating graph neural network training.
\newblock In \emph{International Conference on Machine Learning}, pages 1204--1215, 2021.

\bibitem[Loshchilov and Hutter(2019)]{loshchilov2017_adamw}
Ilya Loshchilov and Frank Hutter.
\newblock Decoupled weight decay regularization.
\newblock In \emph{International Conference on Learning Representations}, 2019.

\bibitem[Choromanski et~al.(2021)Choromanski, Likhosherstov, Dohan, Song, Gane, Sarlos, Hawkins, Davis, Mohiuddin, Kaiser, Belanger, Colwell, and Weller]{choromanski2020_performers}
Krzysztof Choromanski, Valerii Likhosherstov, David Dohan, Xingyou Song, Andreea Gane, Tamas Sarlos, Peter Hawkins, Jared Davis, Afroz Mohiuddin, Lukasz Kaiser, David Belanger, Lucy Colwell, and Adrian Weller.
\newblock Rethinking attention with performers.
\newblock In \emph{International Conference on Learning Representations}, 2021.

\bibitem[Dwivedi et~al.(2022)Dwivedi, Luu, Laurent, Bengio, and Bresson]{dwivedi2021_rwpe}
Vijay~Prakash Dwivedi, Anh~Tuan Luu, Thomas Laurent, Yoshua Bengio, and Xavier Bresson.
\newblock Graph neural networks with learnable structural and positional representations.
\newblock In \emph{International Conference on Learning Representations}, 2022.

\bibitem[Hu et~al.(2020)Hu, Liu, Gomes, Zitnik, Liang, Pande, and Leskovec]{hu2020strategies_GINE}
Weihua Hu, Bowen Liu, Joseph Gomes, Marinka Zitnik, Percy Liang, Vijay Pande, and Jure Leskovec.
\newblock Strategies for pre-training graph neural networks.
\newblock In \emph{International Conference on Learning Representations}, 2020.

\bibitem[Andersen et~al.(2006)Andersen, Chung, and Lang]{andersen2006_acl}
Reid Andersen, Fan Chung, and Kevin Lang.
\newblock Local graph partitioning using {PageRank} vectors.
\newblock In \emph{2006 47th Annual IEEE Symposium on Foundations of Computer Science (FOCS'06)}, pages 475--486. IEEE, 2006.

\end{thebibliography}
